\documentclass[sigconf]{acmart}

\usepackage[utf8]{inputenc} 
\usepackage[T1]{fontenc}    
\usepackage{hyperref}       
\usepackage{url}            
\usepackage{booktabs}       
\usepackage{amsfonts}       
\usepackage{nicefrac}       
\usepackage{microtype}      
\usepackage{subfigure}

\usepackage{amssymb}
\usepackage{multirow}
\newtheorem{defn}{Definition}[section]
\newtheorem{thm}{Theorem}[section]

\newtheorem{remark}{Remark}[section]

\usepackage{graphicx}
\usepackage{float}
\usepackage{mathtools}
\usepackage{wrapfig}
\usepackage{setspace}
\usepackage{bbding}
\usepackage{pifont}
\AtBeginDocument{%
  \providecommand\BibTeX{{%
    \normalfont B\kern-0.5em{\scshape i\kern-0.25em b}\kern-0.8em\TeX}}}

\copyrightyear{2021}
\acmYear{2021} 
\setcopyright{iw3c2w3}
\acmConference[WWW '21]{Proceedings of the Web Conference 2021}{April 19--23, 2021}{Ljubljana, Slovenia} 
\acmBooktitle{Proceedings of the Web Conference 2021 (WWW '21), April 19--23, 2021, Ljubljana, Slovenia}
\acmPrice{}
\acmDOI{10.1145/3442381.3449872}
\acmISBN{978-1-4503-8312-7/21/04}

\begin{document}

\title{Lorentzian Graph Convolutional Networks}



\author{Yiding Zhang}
\email{zyd@bupt.edu.cn}
\affiliation{%
	\institution{Beijing University of Posts and Telecommunications}
	\streetaddress{No.10 Xitucheng Road}
	\state{Beijing}
	\country{China}}

\author{Xiao Wang}
\email{xiaowang@bupt.edu.cn}
\affiliation{%
	\institution{Beijing University of Posts and Telecommunications}
	\streetaddress{No.10 Xitucheng Road}
	\state{Beijing}
	\country{China}}

\author{Chuan Shi}\authornote{Chuan Shi is a corresponding author. E-mail: shichuan@bupt.edu.cn.}
\email{shichuan@bupt.edu.cn}
\affiliation{%
	\institution{Beijing University of Posts and Telecommunications}
	\streetaddress{No.10 Xitucheng Road}
	\state{Beijing}
	\country{China}}

\author{Nian Liu}
\email{nianliu@bupt.edu.cn}
\affiliation{%
	\institution{Beijing University of Posts and Telecommunications}
	\streetaddress{No.10 Xitucheng Road}
	\state{Beijing}
	\country{China}}

\author{Guojie Song}
\email{gjsong@pku.edu.cn}
\affiliation{%
	\institution{Peking University}
	\streetaddress{No.5 Yiheyuan Road}
	\state{Beijing}
	\country{China}}

%
%
%
%
%
%
%

\renewcommand{\shortauthors}{Trovato and Tobin, et al.}

\begin{abstract}
	Graph convolutional networks (GCNs) have received considerable research attention recently.
	Most GCNs learn the node representations in Euclidean geometry, 
	but that could have a high distortion in the case of embedding graphs with scale-free or hierarchical structure.
	Recently, some GCNs are proposed to deal with this problem in non-Euclidean geometry, e.g., hyperbolic geometry.
	Although hyperbolic GCNs achieve promising performance,
	existing hyperbolic graph operations actually cannot rigorously follow the hyperbolic geometry,
	which may limit the ability of hyperbolic geometry and thus hurt the performance of hyperbolic GCNs.
	In this paper,
	we propose a novel hyperbolic GCN named Lorentzian graph convolutional network (LGCN), 
	which rigorously guarantees the learned node features follow the hyperbolic geometry. 
	Specifically, 
	we rebuild the graph operations of hyperbolic GCNs with Lorentzian version,
	e.g., the feature transformation and non-linear activation.
	Also,
	an elegant neighborhood aggregation method is designed based on the centroid of Lorentzian distance.
	Moreover, 
	we prove some proposed graph operations are equivalent in different types of hyperbolic geometry, 
	which fundamentally indicates their correctness.
	Experiments on six datasets
	show that LGCN performs better than the state-of-the-art methods.
	LGCN has lower distortion to learn the representation of tree-likeness graphs compared with existing hyperbolic GCNs.
	We also find that the performance of some hyperbolic GCNs can be improved by simply replacing the graph operations with those we defined in this paper.
\end{abstract}



\maketitle

\section{Introduction}

Graph Convolutional Networks (GCNs) \cite{defferrard2016convolutional,kipf2016semi,hamilton2017inductive}
are powerful deep representation learning methods for graphs.
The current GCNs usually follow a message passing manner, 
where the key steps are feature transformation and neighborhood aggregation.
Specifically,
GCNs leverage feature transformation to transform the features into higher-level features,
and
neighborhood aggregation in GCNs averages the features of its local neighborhood for a given node.
GCNs have aroused considerable attention  \cite{defferrard2016convolutional,kipf2016semi,hamilton2017inductive}
and are widely used in many application areas,
e.g., natural language processing \cite{yao2019graph,linmei2019heterogeneous},
recommendation  \cite{ying2018graph,song2019session} and 
disease prediction \cite{parisot2017spectral,sungmin2017hybrid}.

Most GCNs learn the node features in Euclidean spaces.
However,
some studies find that compared with Euclidean geometry, 
hyperbolic geometry actually can provide more powerful ability to embed graphs with scale-free or hierarchical structure \cite{clauset2008hierarchical,clauset2009power,muscoloni2017machine}.
As a consequence,
several recent efforts begin to define graph operations in hyperbolic spaces  (e.g., feature transformation, neighborhood aggregation),
and propose hyperbolic GCNs in different ways \cite{chami2019hyperbolic, liu2019hyperbolic,zhang2019hyperbolic,bachmann2019constant}.
For instance,
HGCN \cite{chami2019hyperbolic} extends the graph convolution on the hyperboloid manifold of hyperbolic spaces,
while HAT \cite{zhang2019hyperbolic} leverages the Poincar\'e ball manifold to design hyperbolic graph operations.

Despite the promising performance of hyperbolic GCNs, 
existing hyperbolic message passing rules do not rigorously follow hyperbolic geometry,
which may not fully embody the ability of hyperbolic spaces.
Specifically, these hyperbolic GCNs suffer from the following issues:
(1) 
Some hyperbolic graph operations could make node features out of the hyperbolic spaces.
For example,
a critical step of HGCN \cite{chami2019hyperbolic}, the feature transformation, is actually conducted in tangent spaces.
However, it
ignores the constraint of Lorentzian scalar product in tangent spaces,
which leads to the node features deviate from the hyperboloid manifold.
(2) The current hyperbolic neighborhood aggregations do not conform to the same mathematical meanings with Euclidean one,
which could cause a distortion for the learned node features.
Actually, 
the mathematical meanings of Euclidean neighborhood aggregation can be considered as the weighted arithmetic mean or centroid of the representations of node neighbors.
However, 
the neighborhood aggregation in hyperbolic GCN may not obey the similar rules in hyperbolic spaces.
Taking HGCN \cite{chami2019hyperbolic} as an example, 
it aggregates the node features in tangent spaces, 
which can only meet the mathematical meanings in tangent spaces, rather than hyperbolic spaces.
Since we aim to build a hyperbolic GCN,
it is a fundamental requirement to ensure
the basic graph operations rigorously follow the hyperbolic geometry and mathematical meaning, 
so that we can well possess the capability of preserving the graph structure and property in the hyperbolic spaces.
In this paper,
we propose a novel \textit{Lorentzian Graph Convolutional Network} (LGCN), 
which designs a unified framework of graph operations on the hyperboloid model of hyperbolic spaces.
The rigorous hyperbolic graph operations, including feature transformation and non-linearity activation, 
are derived from this framework to ensure the transformed node features follow the hyperbolic geometry.
Also, 
based on the centroid of Lorentzian distance, 
an elegant hyperbolic neighborhood aggregation is proposed to make sure the node features are aggregated to satisfy the mathematical meanings.
Moreover, we theoretically prove that some proposed graph operations are equivalent to those defined in 
another typical hyperbolic geometry, i.e., the Poincar\'e ball model \cite{ganea2018hnn},
so the proposed methods elegantly bridge the relation of these graph operations in different models of hyperbolic spaces,
and also indicates the proposed methods fill the gap of lacking rigorously graph operations on the hyperboloid model.
We conduct extensive experiments to evaluate the performance of LGCN,
well demonstrating the superiority of LGCN in link prediction and node classification tasks,
and LGCN has lower distortion when learning the representation of tree-likeness graphs compared with existing hyperbolic GCNs.
We also find the proposed Lorentzian graph operations can enhance the performance of existing hyperbolic GCN in molecular property prediction task, by simply replacing their operation operations.
\section{Related work}
\subsection{Graph neural networks}
Graph neural networks \cite{gori2005new,scarselli2009graph},
which extend the deep neural network to deal with graph data, 
have achieved great success in solving machine learning problems.
There are two main families of GNNs have been proposed, i.e., spectral methods and spatial methods.
Spectral methods learn node representation via generalizing convolutions to graphs.
Bruna et al. \cite{bruna2013spectral} extended convolution from Euclidean data to arbitrary graph-structured data by finding the corresponding Fourier basis of the given graph.
Defferrard et al. \cite{defferrard2016convolutional} leveraged K-order Chebyshev polynomials to approximate the convolution filter.
Kipf et al. \cite{kipf2016semi} proposed GCN, which utilized a first-order approximation of ChebNet to learn the node representations.
Niepert et al. \cite{niepert2016learning} normalized each node and its neighbors, which served as the receptive field for the convolutional operation.
Wu et al. \cite{wu2019simplifying} proposed simple graph convolution by converting the graph convolution to a linear version.
Moreover,
some researchers defined graph convolutions in the spatial domain.
Li et al. \cite{li2015gated} proposed the gated graph neural network by using the Gate Recurrent Units (GRU) in the propagation step.
Veli{\v{c}}kovi{\'c} et al. \cite{velivckovic2017graph} studied the attention mechanism in GCN to incorporate the attention mechanism into the propagation step.
Chen et al. \cite{chen2018fastgcn} sampled a fix number of nodes for each graph convolutional layer to improve its efficiency.
Ma et al. \cite{ma2020streaming} obtained the sequential information of edges to model the dynamic information as graph evolving.
A comprehensive review can be found in recent surveys \cite{zhang2018deep,wu2020comprehensive}.

\subsection{Hyperbolic graph representation learning}
Recently, node representation learning in hyperbolic spaces has received increasing attention.
Nickel et al. \cite{nickel2017poincare,nickel2018learning} embedded graph into hyperbolic spaces to learn the hierarchical node representation.
Sala et al. \cite{de2018representation} proposed a novel combinatorial embedding approach as well as a approach to Multi-Dimensional Scaling in hyperbolic spaces.
To better modeling hierarchical node representation,
Ganea et al. \cite{ganea2018hyperbolic} and Suzuki et al. \cite{suzuki2019hyperbolic} embedded the directed acyclic graphs into hyperbolic spaces to learn their hierarchical feature representations.
Law et al. \cite{law2019lorentzian} analyzed the relation between hierarchical representations and Lorentzian distance.
Also,
Bala\v{z}evi\'c et al. \cite{balazevic2019multi} analyzed the hierarchical structure in multi-relational graph, and embedded them in hyperbolic spaces.
Moreover, 
some researchers began to study the deep learning in hyperbolic spaces.
Ganea et al. \cite{ganea2018hnn} generalized deep neural models in hyperbolic spaces, such as recurrent neural networks and GRU.
Gulcehre et al. \cite{gulcehre2018hyperbolic} proposed the attention mechanism in hyperbolic spaces.
There are some attempts in hyperbolic GCNs recently.
Liu et al. \cite{liu2019hyperbolic} proposed graph neural networks in hyperbolic spaces which focuses on graph classification problem.
Chami et al. \cite{chami2019hyperbolic} leveraged hyperbolic graph convolution to learn the node representation in hyperboloid model.
Zhang et al. \cite{zhang2019hyperbolic} proposed graph attention network in Poincar\'e ball model to embed some hierarchical and scale-free graphs with low distortion.
Bachmann et al. \cite{bachmann2019constant} also generalized graph convolutional in a non-Euclidean setting.
Although these hyperbolic GCNs have achieved promising results,
we find that some basis properties of GCNs are not well preserved.
so how to design hyperbolic GCNs in a principled manner is still an open question.
The detailed of existing hyperbolic GCNs will be discussed in Section \ref{sec_discuss}.



\section{Preliminaries}

\subsection{Hyperbolic geometry}\label{sec:hyperbolic_geometry}
Hyperbolic geometry is a non-Euclidean geometry with a constant negative curvature.
The hyperboloid model, 
as one typical equivalent model which well describes hyperbolic geometry, 
has been widely used \cite{nickel2018learning,chami2019hyperbolic,liu2019hyperbolic,law2019lorentzian}.
Let  $\mathbf{x},\mathbf{y}\in \mathbb{R}^{n+1}$,
then the \textbf{Lorentzian scalar product} is defined as:
\begin{equation}
	\langle \mathbf{x},\mathbf{y} \rangle_\mathcal{L} := -x_0y_0+\sum_{i=1}^{n}x_i y_i.
\end{equation}
We denote $\mathbb{H}^{n,\beta}$ as the $n$-dimensional hyperboloid manifold with constant negative curvature $-1/\beta$ ($\beta>0$):
\begin{equation}
	\mathbb{H}^{n,\beta}:=\{\mathbf{x}\in \mathbb{R}^{n+1}: \langle \mathbf{x}, \mathbf{x} \rangle_\mathcal{L}=-\beta, x_0>0 \}.
\end{equation}
Also, for $\mathbf{x,y}\in \mathbb{H}^{n,\beta}$,
Lorentzian scalar product satisfies:
\begin{equation}\label{eq_l_inner_eq}
	\langle \mathbf{x, y} \rangle_\mathcal{L}\leq-\beta,\text{  and } \langle \mathbf{x, y} \rangle_\mathcal{L}=-\beta\text{ iff }\mathbf{x}=\mathbf{y}.
\end{equation}
The \textbf{tangent space} at $\mathbf{x}$ is defined as a $n$-dimensional vector space approximating $\mathbb{H}^{n,\beta}$ around $\mathbf{x}$,
\begin{equation}\label{eq:tangent_space}
	\mathcal{T}_\mathbf{x}\mathbb{H}^{n,\beta}:=\{\mathbf{v}\in\mathbb{R}^{n+1}:\langle \mathbf{v}, \mathbf{x} \rangle_\mathcal{L}=0\}.
\end{equation}
Note that Eq. \eqref{eq:tangent_space} has a constraint of Lorentzian scalar product.
Also,
for $\mathbf{v,w}\in \mathcal{T}_\mathbf{x}\mathbb{H}^{n,\beta}$, a Riemannian metric tensor is given as $g_\mathbf{x}^\beta(\mathbf{v,w}):=\langle \mathbf{v,w} \rangle_\mathcal{L} $.
Then the hyperboloid model is defined as the hyperboloid manifold $\mathbb{H}^{n,\beta}$ equipped with the Riemannian metric tensor $g_\mathbf{x}^\beta$.



The mapping between hyperbolic spaces and tangent spaces can be done by \textbf{exponential map}  and \textbf{logarithmic map}.
The exponential map is a map from subset of a tangent space of $\mathbb{H}^{n,\beta}$  (i.e., $\mathcal{T}_\mathbf{x}\mathbb{H}^{n,\beta}$) to $\mathbb{H}^{n,\beta}$ itself.
The logarithmic map is the reverse map that maps back to the tangent space.
For points $\mathbf{x,y}\in \mathbb{H}^{n,\beta}, \mathbf{v}\in \mathcal{T}_\mathbf{x}\mathbb{H}^{n,\beta}$, 
such that $\mathbf{v}\neq\mathbf{0}$ and $\mathbf{x}\neq \mathbf{y}$,
the exponential map $\exp_\mathbf{x}^\beta(\cdot)$ and logarithmic map $\log_\mathbf{x}^\beta
(\cdot)$ are given as follows:
\begin{flalign}\label{eq:exp_map}
	\exp_\mathbf{x}^\beta(\mathbf{v})&=\cosh\!\Big(\frac{\|\mathbf{v}\|_\mathcal{L}}{\sqrt{\beta}}\!\Big)\mathbf{x}+\sqrt{\beta}\sinh\Big(\frac{\|\mathbf{v}\|_\mathcal{L}}{\sqrt{\beta}}\Big)\frac{\mathbf{v}}{\|\mathbf{v}\|_\mathcal{L}},\\
	\log_\mathbf{x}^\beta(\mathbf{y})&=
	d_\mathbb{H}^\beta(\mathbf{x,y})
	\frac{\mathbf{y}+\frac{1}{\beta}\langle\mathbf{x,y}\rangle_\mathcal{L}\mathbf{x}}
	{\|\mathbf{y}+\frac{1}{\beta}\langle\mathbf{x,y}\rangle_\mathcal{L}\mathbf{x}\|_\mathcal{L}},
\end{flalign}
where $\|\mathbf{v}\|_\mathcal{L}=\sqrt{\langle \mathbf{v},\mathbf{v} \rangle_\mathcal{L}}$ denotes Lorentzian norm of $\mathbf{v}$ and $d_\mathbb{H}^{\beta}(\cdot,\cdot)$ denotes the intrinsic distance function between two points $\mathbf{x},\mathbf{y} \in \mathbb{H}^{d, \beta}$, which is given as:
\begin{equation}\label{eq:intrinsic_distance}
	d_\mathbb{H}^{\beta}(\textbf{x}, \textbf{y}) = \sqrt{\beta} \rm{\ arcosh}\big(-{\langle\mathbf{x},\mathbf{y} \rangle_\mathcal{L}}/{\beta} \big).	
\end{equation}



\subsection{Hyperbolic graph convolutional networks}\label{sec:hgcn}

Recently, 
several hyperbolic GCNs have been proposed \cite{chami2019hyperbolic,liu2019hyperbolic,zhang2019hyperbolic,bachmann2019constant}.
Here we use HGCN \cite{chami2019hyperbolic}, which extends Euclidean graph convolution to the hyperboloid model, as a typical example to illustrate the basic framework of hyperbolic GCN.
Let $\mathbf{h}_i^{k,\beta}\in \mathbb{H}^{k,\beta}$
be a $k$-dimensional node feature of node $i$,
$N(i)$ be a set of its neighborhoods with aggregation weight $w_{ij}$,
and $\mathbf{M}$ be a $(d+1)\times(k+1)$ weight matrix.
The message passing rule of HGCN consists of \textit{feature transformation}:
\begin{flalign}
	\mathbf
	{h}^{d,\beta}_i = \exp_\mathbf{0}^\beta(\mathbf{M}\log_\mathbf{0}^\beta(\mathbf{h}_i^{k,\beta})),\label{eq:hgcn_trans}
\end{flalign}
and \textit{neighborhood aggregation}:
\begin{flalign}
	AGG^\beta(\mathbf{h}_i^{d,\beta}) =
	\exp_{\mathbf{h}_i}^\beta \big(\sum_{j\in N(i) \cup \{i\}} w_{ij}
	\log_{\mathbf{h}_i}^\beta(\mathbf{h}_j^{d, \beta})\big).\label{eq:hgcn_agg}
\end{flalign}
As we can see in Eq. \eqref{eq:hgcn_trans}, 
the features are transformed from hyperbolic spaces to tangent spaces via logarithmic map $\log_\mathbf{0}^\beta(\cdot)$.
However, the basic constraint of tangent spaces in Eq. \eqref{eq:tangent_space},  $\langle \mathbf{v}, \mathbf{x} \rangle_\mathcal{L} = 0$, is violated,
since $\langle \mathbf{M}\log_\mathbf{0}^\beta(\mathbf{h}_i^{k,\beta}), \mathbf{0} \rangle_\mathcal{L}\neq 0$,
$\mathbf{0}=(\sqrt{\beta},0,\cdots,0)\in\mathbb{H}^{k,\beta}$.
As a consequence, 
the node features would be out of the hyperbolic spaces after projecting them back to hyperboloid manifold via the exponential map $\exp_\mathbf{0}^\beta$, 
which do not satisfy hyperbolic geometry rigorously.

On the other hand,
in Euclidean spaces,
the node feature 
$\mathbf{h}_i^d\!\in\!\mathbb{R}^d$ aggregates information from its neighborhoods via 
$\!\sum_{\!j\in N(\!i) \cup \!\{\!i\!\}}\!\! w_{\!i\!j}\mathbf{h}_j^{d}$,
which has the following meaning in mathematics:
\begin{remark}\label{remark:mean}
	Given a node,
	the neighborhood aggregation essentially is \textit{the weighted arithmetic mean} for features of its local neighborhoods \cite{wu2019simplifying}.
	Also, the feature of aggregation is the centroid of the neighborhood features in geometry.
\end{remark}
Remark \ref{remark:mean} indicates the mathematical meanings of neighborhood aggregation in Euclidean spaces.
Therefore, 
the neighborhood aggregation in Eq. \eqref{eq:hgcn_agg}
should also follow the same meanings with Euclidean one in hyperbolic spaces.
However, 
we can see that the Eq. \eqref{eq:hgcn_agg} in HGCN only meets these meanings in tangent spaces rather than hyperbolic spaces,
which could cause a distortion for the features.
To sum up,
the above issues indicate existing hyperbolic graph operations do not follow mathematic fundamentally,
which may cause potential untrustworthy problem.





\section{LGCN: Our proposed Model}\label{sec:model}
In order to solve the issues of existing hyperbolic GCNs, 
we propose LGCN,
which designs graph operations to guarantee the mathematical meanings in hyperbolic spaces.
Specifically,
LGCN first maps the input node features into hyperbolic spaces and then conducts feature transformation via a delicately designed Lorentzian matrix-vector multiplication.
Also,
the centroid based
Lorentzian aggregation is proposed to aggregate features,
and the aggregation weights are learned by a self attention mechanism.
Moreover,
Lorentzian pointwise non-linear activation is followed to obtain the output node features.
Note that the curvature of a hyperbolic space (i.e., $-1/\beta$) is also a trainable parameter for LGCN.
Despite the same expressive power, 
adjusting curvature of LGCN is important in practice due to factors of limited machine precision and normalization.
The details of LGCN are introduced in the following.

\subsection{Mapping feature with different curvature}
The input node features of LGCN could live in the Euclidean spaces or hyperbolic spaces.
For $k$-dimensional input features,
we denote them as $\mathbf{h}^{k,E}\in\mathbb{R}^k$ ($^E$ indicates Euclidean spaces) and $\mathbf{h}^{k,\beta'}\in\mathbb{H}^{k,\beta'}$, respectively.
If original features live in Euclidean spaces, 
we need to map them into hyperbolic spaces.
We assume that the input features $\mathbf{h}^{k,E}$ live in the tangent space of $\mathbb{H}^{k,\beta}$ at its origin $\mathbf{0}=(\sqrt{\beta},0,\cdots,0)\in\mathbb{H}^{k,\beta}$,
i.e., $\mathcal{T}_\mathbf{0}\mathbb{H}^{k,\beta}$.
A ``0'' element is added at the first coordinate of  $\mathbf{h}^{k,E}$ to satisfy the constraint
$\langle (0, \mathbf{h}^{k,E}), \mathbf{0} \rangle_\mathcal{L}=0$
in Eq. \eqref{eq:tangent_space}.
Thus, the input feature $\mathbf{h}^{k,E}\in\mathbb{R}^k$ can be mapped to the hyperbolic spaces via exponential map:
\begin{equation}\label{eq:input_map_euc}
	\mathbf{h}^{k,\beta} 
	= \exp_\mathbf{0}^\beta\big((0, \mathbf{h}^{k,E})\big).
\end{equation}
If
the input features $\mathbf{h}^{k,\beta'}$ live in a hyperbolic space (e.g., the output of previous LGCN layer),
whose curvature $-1/\beta'$ might be different with the curvature of current hyperboloid model.
We can transform it into the hyperboloid model with a specific curvature $-1/\beta$:
\begin{equation}
	\mathbf{h}^{k,\beta} = \exp_\mathbf{0}^\beta(\log_\mathbf{0}^{\beta'}(\mathbf{h}^{k,\beta'})).
\end{equation}
%

\subsection{Lorentzian feature transformation}

Hyperbolic spaces are not vector spaces,
which means the operations in Euclidean spaces cannot be applied in hyperbolic spaces.
To ensure the transformed features satisfy the hyperbolic geometry,
it is crucial to define some canonical transformations in the hyperboloid model, so we define:
\begin{defn}[Lorentzian version]\label{def:l_version}
	For $f\!:\mathbb{R}^n\!\rightarrow\!\mathbb{R}^m\ (n,m>2)$ and two points $\mathbf{\mathbf{x}}=(x_0,\cdots,x_n)\in\mathbb{H}^{n,\beta}$, $\mathbf{v}=(v_0,\cdots,v_n)\in \mathcal{T}_\mathbf{0}\mathbb{H}^{n,\beta}$,  
	we define the Lorentzian version of $f$ as the map $\mathbb{H}^{n,\beta}\rightarrow\mathbb{H}^{m,\beta}$ by:
	\begin{equation}
		f^{\otimes^\beta}(\mathbf{x}):=\exp_\mathbf{0}^\beta(\hat{f}(\log_\mathbf{0}^\beta(\mathbf{x}))),
		\hat{f}(\mathbf{v}):=(0, f(v_1,\cdots,v_n)),
	\end{equation}
	where $\exp_\mathbf{0}^\beta:\mathcal{T}_\mathbf{0}\mathbb{H}^{n,\beta}\rightarrow\mathbb{H}^{m,\beta}$
	and $\log_\mathbf{0}^\beta:\mathbb{H}^{n,\beta}\rightarrow\mathcal{T}_\mathbf{0}\mathbb{H}^{m,\beta}$.
\end{defn}
Lorentzian version leverages logarithmic and exponential map to project the features between hyperbolic spaces and tangent spaces.
As the tangent spaces are vector spaces and isomorphic to $\mathbb{R}^n$, the Euclidean transformations can be applied to the tangent spaces.
Moreover,
given a point $\mathbf{v}=(v_0,\cdots,v_n)\in\mathcal{T}_\mathbf{0}\mathbb{H}^{n,\beta}$,
existing methods \cite{chami2019hyperbolic,liu2019hyperbolic} directly apply the Euclidean transformations on all coordinates $(v_0,\cdots, v_n)$ in tangent spaces.
Different from these methods,
Lorentzian version only leverages the Euclidean transformations on the last $n$ coordinates $(v_1,\cdots,v_n)$ in tangent spaces,
and
the first coordinate $(v_0)$ is set as ``$0$'' to satisfy the constraint in Eq. \eqref{eq:tangent_space}.
Thus, this operation can make sure the transformed features rigorously follow the hyperbolic geometry.

In order to apply \textbf{linear transformation} on the hyperboloid model,
following Lorentzian version, 
the Lorentzian matrix-vector multiplication can be derived:
\begin{defn}[Lorentzian matrix-vector multiplication]
	If $\mathbf{M}:\mathbb{R}^n\rightarrow\mathbb{R}^m$ is a linear map with matrix representation, 
	given two points $\mathbf{\mathbf{x}}=(x_0,\cdots,x_n)\in\mathbb{H}^{n,\beta}$,
	$\mathbf{v}=(v_0,\cdots,v_n)\in \mathcal{T}_\mathbf{0}\mathbb{H}^{n,\beta}$, 
	we have:
	\begin{equation}
		\mathbf{M}^{\otimes^\beta}(\mathbf{x})=\exp_\mathbf{0}^\beta(\hat{\mathbf{M}}(\log_\mathbf{0}^\beta(\mathbf{x}))),\\
		\hat{\mathbf{M}}(\mathbf{v})=(0, \mathbf{M}(v_1,\cdots,v_n)).
	\end{equation}
	Let $\mathbf{M}$ be a $m\times n$ matrix, $\mathbf{M}'$ be a $l\times m$ matrix, $\mathbf{x}\in\mathbb{H}^{n,\beta}$,
	$\mathbf{M}{\otimes^\beta}\mathbf{x}:=\mathbf{M}^{\otimes^\beta}(\mathbf{x})$,
	we have matrix associativity as: $(\mathbf{M}'\mathbf{M}){\otimes^\beta}\mathbf{x}
	= \mathbf{M}'\otimes^\beta(\mathbf{M}{\otimes^\beta}\mathbf{x}) $.
\end{defn}
%
A key difference between Lorentzian matrix-vector multiplication and other matrix-vector multiplications on the hyperboloid model \cite{chami2019hyperbolic,liu2019hyperbolic} is the size of the matrix $\mathbf{M}$.
Assuming  a $n$-dimensional feature needs to be transformed into a $m$-dimensional feature.
Naturally, the size of matrix $\mathbf{M}$ should be $m\times n$,
which is satisfied by Lorentzian matrix-vector multiplication. 
However,
the size of matrix $\mathbf{M}$ is $(m+1)\times(n+1)$ for other methods \cite{chami2019hyperbolic,liu2019hyperbolic} (as shown in Eq. \eqref{eq:hgcn_trans}),
which leads to the constraint of tangent spaces  cannot be satisfied, i.e., 
$\langle \mathbf{M}\log_\mathbf{0}^\beta(\mathbf{h}_i^{k,\beta}), \mathbf{0} \rangle_\mathcal{L}\neq 0$
in Eq. \eqref{eq:tangent_space}, so
the transformed features would be out of the hyperbolic spaces.
Moreover,
the Lorentzian matrix vector multiplication has the following property:
\begin{thm}\label{thm_mv}
	Given a point in hyperbolic space,
	which is represented by $\mathbf{x}^{n,\beta}\in\mathbb{H}^{n,\beta}$ using hyperboloid model or $\mathbf{x}^{n,\alpha}\in\mathbb{D}^{n,\alpha}$ using Poincar\'e ball model \cite{ganea2018hnn}, respectively.
	Let $\mathbf{M}$ be a $m\times n$ matrix,
	Lorentzian matrix-vector multiplication $\mathbf{M}{\otimes^\beta}\mathbf{x}^{n,\beta}$ used in hyperboloid model is equivalent to M\"obius matrix-vector multiplication $\mathbf{M}{\otimes^\alpha}\mathbf{x}^{n,\alpha}$ used in Poincar\'e ball model.
\end{thm}
The proof is in Appendix \ref{sec:app_proof3.1}.
This property elegantly bridges the relation between the hyperboloid model and Poincar\'e ball model w.r.t. matrix-vector multiplication.
We use the Lorentzian matrix-vector multiplication to conduct feature transformation on the hyperboloid model as:
\begin{equation}
	\mathbf{h}^{d,\beta} = \mathbf{M}{\otimes^\beta}\mathbf{h}^{k, \beta}.
\end{equation}

\begin{figure*}[t]
	\centering
	\subfigure[\scriptsize Euclidean aggregation in Euclidean spaces]{
		\includegraphics[width=0.31\textwidth]{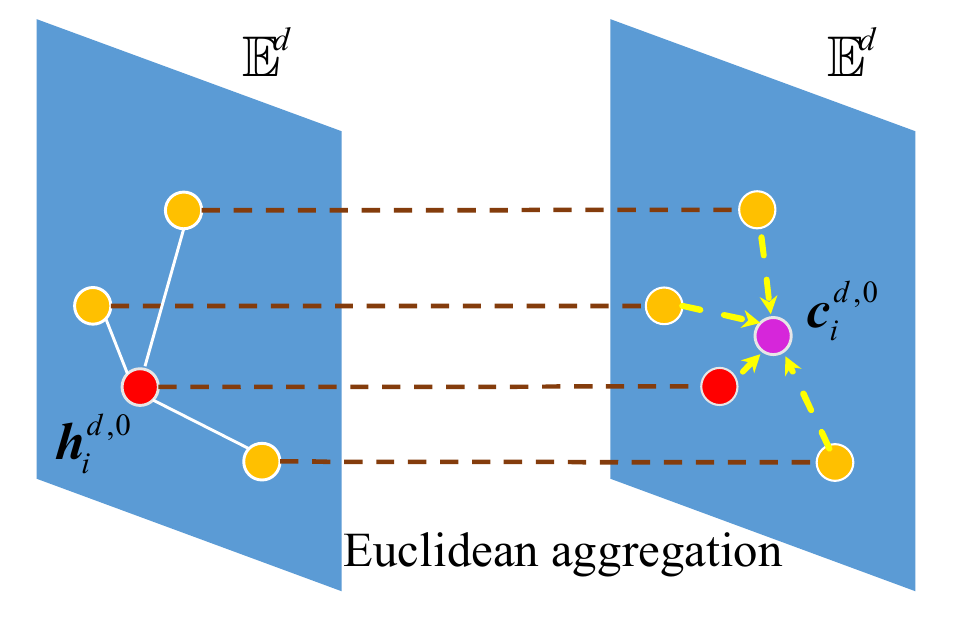}
		\label{fig_agg_euc}
	}
	\subfigure[\scriptsize Hyperbolic aggregation on hyperboloid model via centroid]{
		\includegraphics[width=0.54\textwidth]{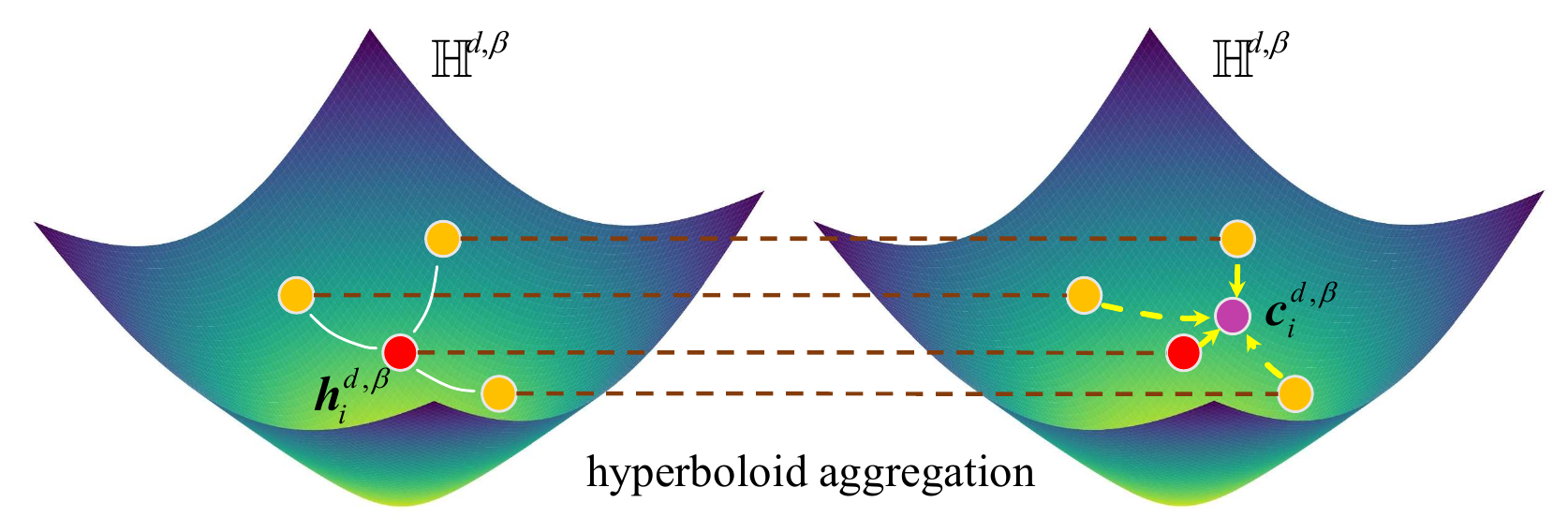}
		\label{fig_agg_hyp}
	}
	\subfigure[\scriptsize Hyperbolic aggregation in tangent spaces]{
		\includegraphics[width=0.9\textwidth]{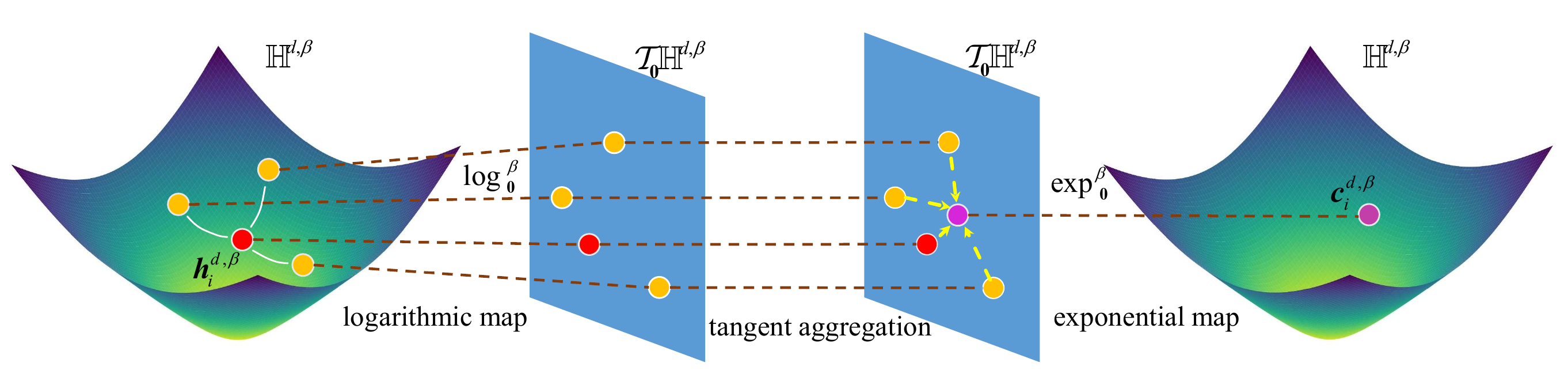}
		\label{fig_agg_tangent}
	}
	\caption{Three types of neighborhood aggregation. The three types of aggregation can be considered as computing centroids in Euclidean spaces, hyperbolic spaces, tangent spaces, respectively.}
	\label{fig_agg}
\end{figure*}

\subsection{Lorentzian neighborhood aggregation}

As in Remark \ref{remark:mean}, in Euclidean spaces,
the neighborhood aggregation is to compute the weight arithmetic mean or centroid (also called center of mass) of its neighborhood features 
(see Fig. \ref{fig_agg_euc}).
Therefore,
we aim to aggregate neighborhood features in hyperbolic spaces to follow these meanings.
Fr\'echet mean
\cite{frechet1948elements,karcher1987riemannian,karcher2014riemannian}
provides a feasible way to compute the centroid in Riemannian manifold.
Also, the arithmetic mean can be interpreted as a kind of Fr\'echet mean.
Thus, Fr\'echet mean meets the meanings of neighborhood aggregation.
The main idea of Fr\'echet mean is to minimize an expectation of (squared) distances with a set of points.
However,
Fr\'echet mean does not have a closed form solution w.r.t. the intrinsic distance $d_{\mathbb{H}}^\beta$ in hyperbolic spaces,
and it has to be inefficiently computed by gradient descent.
Therefore, we propose an elegant neighborhood aggregation method based on the centroid of the squared Lorentzian distance,
which can well balance the mathematical meanings and efficiency:

\begin{thm}[Lorentzian aggregation via centroid of squ- ared Lorentzian distance]\label{thm_agg}
	For a node feature $\mathbf{h}_i^{d,\beta}\in\mathbb{H}^{d,\beta}$, a set of its neighborhoods $N(i)$ with aggregation weights $w_{ij}>0$,
	the  neighborhood aggregation consists in the centroid $\mathbf{c}^{d,\beta}$ of nodes, which minimizes the problem:
	\begin{equation}\label{eq_except_agg}
		\arg\min _{\mathbf{c}^{d,\beta}\in\mathbb{H}^{d,\beta}}\sum_{j\in N(i) \cup \{i\}} w_{ij}d_{\mathcal{L}}^2(\mathbf{h}_j^{d,\beta}, \mathbf{c}^{d,\beta}),
	\end{equation}
	where $d_\mathcal{L}^2(\cdot,\cdot)$ denotes squared Lorentzian distance,
	and this problem has closed form solution:
	\begin{equation}\label{eq:agg_hyperboloid}
		\mathbf{c}^{d,\beta}=\sqrt{\beta}\frac{\sum_{{j\in N(i) \cup \{i\}}}w_{ij}\mathbf{h}_j^{d,\beta}}{|\|\sum_{{j\in N(i) \cup \{i\}}}w_{ij}\mathbf{h}_j^{d,\beta}\|_{\mathcal{L}}|}.
	\end{equation}
\end{thm}

The proof is given in Appendix \ref{sec:app_proof3.2}.
For points $\mathbf{x}^{n,\beta}, \mathbf{y}^{n,\beta}\in\mathbb{H}^{n,\beta}$, the squared Lorentzian distance is defined as \cite{ratcliffe1994foundations}:
\begin{equation}\label{eq_lorentz_dis}
	d_\mathcal{L}^2(\mathbf{x}^{n,\beta},\mathbf{y}^{n,\beta})=-2\beta-2\langle \mathbf{x}^{n,\beta},\mathbf{y}^{n,\beta}\rangle_\mathcal{L}.
\end{equation}

Fig. \ref{fig_agg_hyp} illustrates Lorentzian aggregation via centroid.
Similar to Fr\'echet/Karcher means,
the node features computed by Lorentzian aggregation are the minimum of an expectation of squared Lorentzian distance.
Also,
the features of aggregation in Lorentzian neighborhood aggregation are the centroids in the hyperboloid model
in geometry \cite{ratcliffe1994foundations,law2019lorentzian}.
On the other hand,
some hyperbolic GCNs \cite{chami2019hyperbolic,liu2019hyperbolic,zhang2019hyperbolic} aggregate neighborhoods in tangent spaces (as shown in Fig. \ref{fig_agg_tangent}),
that can only be regarded as centroid or arithmetic mean in the tangent spaces, rather than hyperbolic spaces.
Thus \textit{Lorentzian aggregation via centroid of squared Lorentzian distance}
is a promising method,
which satisfies more elegant mathematical meanings compared to other hyperbolic GCNs.

As shown in Eq. \eqref{eq:agg_hyperboloid}, there is an aggregation weight $w_{ij}$ indicating the importance of neighborhoods for a center node.
Here we propose a self-attention mechanism to learn the aggregation weights $w_{ij}$.
For two node features $\mathbf{h}^{d,\beta}_{i},\mathbf{h}^{d,\beta}_{i}\in\mathbb{H}^{d,\beta}$,
the attention coefficient $\mu_{ij}$, which indicates the importance of node $j$ to node
$i$, can be computed as:
\begin{equation}
	\mu_{ij} = ATT(\mathbf{h}^{d,\beta}_{i},\mathbf{h}^{d,\beta}_j,\mathbf{M}_{att}),
\end{equation}
where $ATT(\cdot)$ indicates the function of computing the attention coefficient
and the $d\times d$ matrix $\mathbf{M}_{att}$ is to transform the node features into attention-based ones.
Considering a large attention coefficient $\mu_{ij}$ represents a high similarity
of nodes $j$ and $i$, 
we define $ATT(\cdot)$ based on squared Lorentzian distance,  as
\begin{equation}
	\mu_{ij} = -d_\mathcal{L}^2(\mathbf{M}_{att}\otimes^\beta\mathbf{h}^{d,\beta}_{i}, \mathbf{M}_{att}\otimes^\beta\mathbf{h}^{d,\beta}_j).
\end{equation}
For all the neighbors $N(i)$ of node $i$ (including itself), 
we normalize them using the softmax function to compute the aggregation weight:
\begin{equation}
	w_{ij} = \frac{\exp(\mu_{ij})}{\sum_{{t\in N(i) \cup \{i\}}}\exp(\mu_{it})}.
\end{equation}



\subsection{Lorentzian pointwise non-linear activation}
Non-linear activation is an indispensable part of GCNs.
Similar to feature transformation,
existing non-linear activations on the hyperboloid model \cite{chami2019hyperbolic} also make features out of the hyperboloid model.
Here, we derive the Lorentzian pointwise non-linear activation following the Lorentzian version:

\begin{defn}[Lorentzian pointwise non-linear activation]
	If $\sigma:\mathbb{R}^n\rightarrow\mathbb{R}^n$ is a pointwise non-linearity map,
	given two points 
	$\mathbf{\mathbf{x}}=(x_0,\cdots,x_n)\in\mathbb{H}^{n,\beta}$ and
	$\mathbf{v}=(v_0,\cdots,v_n)\in \mathcal{T}_\mathbf{0}\mathbb{H}^{n,\beta}$, 
	the Lorentzian version $\sigma^{\otimes^\beta}$ is:
	\begin{equation}
		\sigma^{\otimes^\beta}(\mathbf{x})=\exp_\mathbf{0}^\beta(\hat{\sigma}^{\otimes^\beta}(\log_\mathbf{0}^\beta(\mathbf{x}))),
		\hat{\sigma}^{\otimes^\beta}(\mathbf{v})=(0, \sigma(v_1),\cdots,\sigma(v_n))).
	\end{equation}
\end{defn}
The Lorentzian pointwise non-linear activation not only ensures the transformed features still live in the hyperbolic spaces, but also has the following property.
\begin{thm}\label{thm_nonlinear}
	Given a point in hyperbolic space,
	it is modeled by $\mathbf{x}^{n,\beta}\in\mathbb{H}^{n,\beta}$ using hyperboloid model and $\mathbf{x}^{n,\alpha}\in\mathbb{D}^{n,\alpha}$ using Poincar\'e ball model, respectively.
	Lorentzian pointwise non-linearity $\sigma^{\otimes^\beta}(\mathbf{x}^{n,\beta})$ in the hyperboloid model is equivalent to M\"obius pointwise non-linearity $\sigma^{\otimes^\alpha}(\mathbf{x}^{n,\alpha})$ in the Poincar\'e ball model \cite{ganea2018hnn},
	when $\sigma(\cdot)$ indicates some specific non-linear activation, e.g., Relu, leaklyRelu.
\end{thm}
The proof is in Appendix \ref{sec:app_proof3.3}.
This property also bridges the pointwise non-linearity in the two models.
Following the Lorentzian pointwise non-linear activation,
the output of the LGCN layer is: 
\begin{equation}
	\mathbf{u}^{d,\beta}= \sigma^{\otimes^\beta}(\mathbf{c}^{d,\beta}),
\end{equation}
which can be used to downstream tasks, e.g., link prediction and node classification.


\subsection{Discussion on related works}\label{sec_discuss}
\begin{table}
	\caption{Hyperbolic graph operations.}
	\label{table_discuess_operation}
	\centering
	\resizebox{0.4\textwidth}{!}{
		\begin{tabular}{cc|ccc}
			\toprule
			Method	& Manifold 	& inside$_f$ 	& inside$_n$ & except-agg\\
			\midrule
			HGNN$_P$&			&	\Checkmark	& \Checkmark & \ding{55}\\
			HAT		&Poincar\'e ball&	\Checkmark	& \Checkmark & \ding{55}\\
			$\kappa$GCN	&			&	\Checkmark	& \Checkmark & \ding{55}\\
			\midrule
			HGNN$_H$&			&	\ding{55}	&	-		 &	\ding{55}\\
			HGCN	&Hyperboloid&	\ding{55}	& \ding{55}	 &	\ding{55}\\
			LGCN	&			&\Checkmark		& \Checkmark &	\Checkmark\\
			\bottomrule
		\end{tabular}
	}
\end{table}

\subsubsection{Hyperbolic graph operations}
We compare LGCN with some existing hyperbolic GCNs regarding the properties of graph operations.
A rigorous hyperbolic graph operation should make sure the features still live in the hyperbolic spaces after applying the graph operation.
We analyze this property about feature transformation and pointwise non-linearity activation, denoted as \textit{inside$_f$} and \textit{inside$_n$}, respectively.
Also,
as mentioned in Theorem \ref{thm_agg},
similar with Fr\'echet means,
the neighborhood aggregation to minimize an expectation of distances could better satisfy the mathematical meanings, 
and this property is denoted as \textit{expect-agg}.

The current hyperbolic GCNs can be classified into two classes: 
Poincar\'e ball GCNs, including HGNN$_{P}$ \cite{liu2019hyperbolic}, HAT \cite{zhang2019hyperbolic} and $\kappa$GCN \cite{bachmann2019constant};
Hyperboloid GCNs, i.e.,  HGCN \cite{chami2019hyperbolic}, HGNN$_{H}$ \cite{liu2019hyperbolic} and LGCN.
We summarize the properties of graph operations of these hyperbolic GCNs in Table \ref{table_discuess_operation}.
It can be seen that:
(1)
The existing hyperbolic GCNs do not have all of the three properties except LGCN.
More importantly,
none of the existing hyperbolic neighborhood aggregation satisfy \textit{expect-agg}.
(2) 
All the Poincar\'e ball GCNs satisfy \textit{inside$_f$} and \textit{inside$_n$},
while existing hyperboloid GCNs cannot make sure these properties.
That is because they do not consider the constrain of tangent spaces and the transformed features will be outside of the hyperboloid.
Note that because of lacking non-linear activation on the hyperboloid model,
HGNN$_H$ avoids this problem by conducting non-linear activation on the Poincar\'e ball,
which is implemented via projecting node representations between the Poincar\'e ball and hyperboloid model.
That brings extra computing cost, and also indicates a principle definition of graph operations is needed for the hyperboloid model.
On the other hand,
LGCN fills this gap of lacking rigorously graph operations on the hyperboloid model to ensure the features can be transformed following hyperbolic geometry.
(3)
Only LGCN satisfies \textit{expect-agg}.
Most hyperbolic GCNs \cite{chami2019hyperbolic,liu2019hyperbolic,zhang2019hyperbolic}
leverage aggregation in the tangent spaces (as shown in Fig. \ref{fig_agg_tangent}), which satisfies \textit{expect-agg} in the tangent spaces, instead of the hyperbolic spaces.

\begin{table}
	\caption{Hyperbolic centroids.}
	\label{table_discuess_centroid}
	\centering
	\resizebox{0.46\textwidth}{!}{
		\begin{tabular}{cc|ccc}
			\toprule
			Centroid			& Manifold 	& sum-dis & closed-form  & literature\\
			\midrule
			Fr\'echet mean \cite{frechet1948elements,karcher1987riemannian}		& -	&	\Checkmark	& \ding{55}	 & \cite{sala2018representation,wilson2018gradient,lou2020differentiating}\\
			Einstein gyromidpoint \cite{ungar2005analytic}& Klein ball &	\ding{55}	& \Checkmark &  \cite{gulcehre2018hyperbolic}\\
			M\"obius gyromidpoint \cite{ungar2010barycentric}& Poincar\'e ball&	\ding{55}	& \Checkmark & \cite{bachmann2019constant} \\
			Lorentzian centroid \cite{ratcliffe1994foundations}&Hypreboloid&	\Checkmark	& \Checkmark & \cite{law2019lorentzian}\\
			\bottomrule
		\end{tabular}
	}
\end{table}
\subsubsection{Hyperbolic centroids}
There are some works exploit hyperbolic centroids.
Actually,
the centroid in metric spaces is to find a point which minimizes the sum of squared distance w.r.t. given points \cite{frechet1948elements},
and we denote this property as \textit{sum-dis}.
Also,
the efficiency of computing centroid is important,
so we concern whether a centroid has a closed-form solution, and this property is denoted as \textit{closed-form}.

We summarize hyperbolic centroids as well as some related works in Table \ref{table_discuess_centroid}.
Fr\'echet mean  \cite{frechet1948elements,karcher1987riemannian} is a generalization of centroids to metric spaces by minimizing the sum of squared distance.
Some works \cite{sala2018representation,wilson2018gradient,lou2020differentiating} use Fr\'echet mean in hyperbolic spaces,
which do not have closed-form solution,
so they have to compute them via gradient descent.
Moreover,
Einstein \cite{ungar2005analytic} and M\"obius gyromidpoint \cite{ungar2010barycentric}
are centroids with close-form solution for two different kind of hyperbolic geometry,
i.e., the Klein ball and Poincar\'e ball model, respectively.
Some researchers \cite{gulcehre2018hyperbolic,bachmann2019constant} exploit
Einstein/M\"obius gyromidpoint in representation learning problem.
One limitation of Einstein and M\"obius gyromidipoint is they cannot be seen as minimizing the sum of squared distances.
Furthermore, 
Lorentzian centroid \cite{ratcliffe1994foundations} is the centroid for the hyperboloid model,
which can be seen as a sum of squared distance and has closed-form solution.
The relations between Lorentzian centroid and hierarchical structure data are analyzed in representations learning problem \cite{law2019lorentzian}.
To sum up,
only Lorentzian centroid satisfies the two properties, and we are the first one to leverage it in hyperbolic GCN.





\section{Experiments}

\subsection{Experimental setup}

\begin{table}[t]
	\caption{Dataset statistic}
	\label{table_dataset}
	\centering
	\begin{tabular}{ccccc}
		\toprule
		Dataset	&	Nodes	&	Edges	&	Label	&	Node features\\
		\midrule
		Cora	&	2708	&	5429	&	7		&	1433\\
		Citeseer&	3327	&	4732	&	6		&	3703\\
		Pubmed	&	19717	&	44338	&	3		&	500\\
		Amazon	&	13381	&	245778	&	10		&	767\\
		USA		&	1190	&	13599	&	4		&	-	\\
		Disease	&	1044	&	1043	&	2		&	1000\\
		\bottomrule
	\end{tabular}
\end{table}

\begin{table*}
	\caption{AUC (\%) for link prediction task. The best results are marked by bold numbers. }
	\centering
	\label{table_linkpred}
	\resizebox{0.9\textwidth}{!}{
		\begin{tabular}{cc| cc ccc ccc c}
			\toprule
			Dataset 	& dimension & deepwalk & poincar\'eEmb & GraphSage & GCN & GAT & HGCN & $\kappa$GCN & HAT & LGCN \\ \midrule
			& 8  & 57.3±1.0 & 67.9±1.1 & 65.4±1.4 & 76.9±0.8 & 73.5±0.8 & 84.1±0.7 & 85.3±0.8 & 83.9±0.7 & \textbf{89.2±0.7} \\
			Disease 		& 16 & 55.2±1.7 & 70.9±1.0 & 68.1±1.0 & 78.2±0.7 & 73.8±0.6 & 91.2±0.6 & 92.0±0.5 & 91.8±0.5 & \textbf{96.6±0.6} \\ 
			$\delta_{avg}=0.00$ & 32 & 49.1±1.3 & 75.1±0.7 & 69.5±0.6 & 78.7±0.5 & 75.7±0.3 & 91.8±0.3 & 94.5±0.6 & 92.3±0.5 & \textbf{96.3±0.5} \\ 
			& 64 & 47.3±0.1 & 76.3±0.3 & 70.1±0.7 & 79.8±0.5 & 77.9±0.3 & 92.7±0.4 & 95.1±0.6 & 93.4±0.4 & \textbf{96.8±0.4} \\ \midrule
			& 8  & 91.5±0.1 & 92.3±0.2 & 82.4±0.8 & 89.0±0.6 & 89.6±0.9 & 91.6±0.8 & 92.0±0.6 & 92.7±0.8 & \textbf{95.3±0.2} \\ 
			USA 		& 16 & 92.3±0.0 & 93.6±0.2 & 84.4±1.0 & 90.2±0.5 & 91.1±0.5 & 93.4±0.3 & 93.3±0.6 & 93.6±0.6 & \textbf{96.3±0.2} \\ 
			$\delta_{avg}=0.16$ & 32 & 92.5±0.1 & 94.5±0.1 & 86.6±0.8 & 90.7±0.5 & 91.7±0.5 & 93.9±0.2 & 93.2±0.3 & 94.2±0.6 & \textbf{96.5±0.1} \\ 
			& 64 & 92.5±0.1 & 95.5±0.1 & 89.3±0.3 & 91.2±0.3 & 93.3±0.4 & 94.2±0.2 & 94.1±0.5 & 94.6±0.6 & \textbf{96.4±0.2} \\ \midrule
			& 8  & 96.1±0.0 & 95.1±0.4 & 90.4±0.3 & 91.1±0.6 & 91.3±0.6 & 93.5±0.6 & 92.5±0.7 & 94.8±0.8 & \textbf{96.4±1.1} \\ 
			Amazon	 	& 16 & 96.6±0.0 & 96.7±0.3 & 90.8±0.5 & 92.8±0.8 & 92.8±0.9 & 96.3±0.9 & 94.8±0.5 & 96.9±1.0 & \textbf{97.3±0.8} \\
			$\delta_{avg}=0.20$ & 32 & 96.4±0.0 & 96.7±0.1 & 92.7±0.2 & 93.3±0.9 & 95.1±0.5 & 97.2±0.8 & 94.7±0.5 & 97.1±0.7 & \textbf{97.5±0.3} \\
			& 64 & 95.9±0.0 & 97.2±0.1 & 93.4±0.4 & 94.6±0.8 & 96.2±0.2 & 97.1±0.7 & 95.3±0.2 & 97.3±0.6 & \textbf{97.6±0.5} \\ \midrule
			& 8  & 86.9±0.1 & 84.5±0.7 & 87.4±0.4 & 87.8±0.9 & 87.4±1.0 & 91.4±0.5 & 90.8±0.6 & 91.1±0.4 & \textbf{92.0±0.5} \\
			Cora 		& 16 & 85.3±0.8 & 85.8±0.8 & 88.4±0.6 & 90.6±0.7 & 93.2±0.4 & 93.1±0.4 & 92.6±0.4 & 93.0±0.3 & \textbf{93.6±0.4} \\
			$\delta_{avg}=0.35$ & 32 & 82.3±0.4 & 86.5±0.6 & 88.8±0.4 & 92.0±0.6 & 93.6±0.3 & 93.3±0.3 & 92.8±0.5 & 93.1±0.3 & \textbf{94.0±0.4} \\
			& 64 & 81.6±0.4 & 86.7±0.5 & 90.0±0.1 & 92.8±0.4 & 93.5±0.3 & 93.5±0.2 & 93.0±0.7 & 93.3±0.3 & \textbf{94.4±0.2} \\ \midrule
			& 8  & 81.1±0.1 & 83.3±0.5 & 86.1±1.1 & 86.8±0.7 & 87.0±0.8 & 94.6±0.2 & 93.5±0.5 & 94.4±0.3 & \textbf{95.4±0.2} \\
			Pubmed 		& 16 & 81.2±0.1 & 85.1±0.5 & 87.1±0.4 & 90.9±0.6 & 91.6±0.3 & 96.1±0.2 & 94.9±0.3 & 96.2±0.3 & \textbf{96.6±0.1} \\
			$\delta_{avg}=0.36$ & 32 & 76.4±0.1 & 86.5±0.1 & 88.2±0.5 & 93.2±0.5 & 93.6±0.2 & 96.2±0.2 & 95.0±0.3 & 96.3±0.2 & \textbf{96.8±0.1} \\
			& 64 & 75.3±0.1 & 87.4±0.1 & 88.8±0.5 & 93.6±0.4 & 94.6±0.2 & 96.5±0.2 & 94.9±0.5 & 96.5±0.1 & \textbf{96.9±0.0} \\ \midrule
			& 8  & 80.7±0.3 & 79.2±1.0 & 85.3±1.6 & 90.3±1.2 & 89.5±0.9 & 93.2±0.5 & 92.6±0.7 & 93.1±0.3 & \textbf{93.9±0.6} \\
			Citeseer 	& 16 & 78.5±0.5 & 79.7±0.7 & 87.1±0.9 & 92.9±0.7 & 92.2±0.7 & 94.3±0.4 & 93.8±0.4 & 93.6±0.5 & \textbf{95.4±0.5} \\
			$\delta_{avg}=0.46$ & 32 & 73.1±0.4 & 79.8±0.6 & 87.3±0.4 & 94.3±0.6 & 93.4±0.4 & 94.7±0.3 & 93.5±0.5 & 94.2±0.5 & \textbf{95.8±0.3} \\
			& 64 & 72.3±0.3 & 79.6±0.6 & 88.1±0.4 & 95.4±0.5 & 94.4±0.3 & 94.8±0.3 & 93.8±0.5 & 94.3±0.2 & \textbf{96.4±0.2} \\ \bottomrule
		\end{tabular}
	}
\end{table*}

\subsubsection{Dataset}
We utilize six datasets in our experiments: Cora, Citeseer, Pubmed, \cite{yang2016revisiting}
Amazon \cite{McAuleyTSH15,shchur2018pitfalls},
USA \cite{ribeiro2017struc2vec},
and Disease \cite{chami2019hyperbolic}. 
Cora, Citeseer and Pubmed are citation networks where nodes represent scientific papers, and edges are citations between them.
The Amazon is a co-purchase graph,
where nodes represent goods and edges indicate that two goods are frequently bought together.
The USA is a air-traffic network, 
and the nodes corresponding to different airports.
We use one-hot encoding nodes in the USA dataset as the node features.
The Disease dataset is a graph with tree structure,
where node features indicate the
susceptibility to the disease.
The details of data statistics are shown in the Table \ref{table_dataset}.
We compute $\delta_{avg}$-hyperbolicity \cite{albert2014topological} to quantify the tree-likeliness of these datasets.
A low $\delta_{avg}$-hyperbolicity of a graph indicates 
that it has an underlying hyperbolic geometry.
The details about $\delta_{avg}$-hyperbolicity are shown in Appendix \ref{sec:app_hyperbolicity}.

\subsubsection{Baselines}
We compare our method with the following state-of-the-art methods:
(1) A Euclidean network embedding model i.e., DeepWalk \cite{perozzi2014deepwalk} and a hyperbolic network embedding model i.e., Poincar\'eEmb \cite{nickel2017poincare};
(2) Euclidean GCNs i.e., GraphSage \cite{hamilton2017inductive}, GCN \cite{kipf2016semi}, GAT \cite{velivckovic2017graph};
(3) Hyperbolic GCNs
i.e., HGCN \cite{chami2019hyperbolic}, $\kappa$GCN \cite{bachmann2019constant}\footnote{
	We only consider the $\kappa$GCN in the hyperbolic setting since we focus on hyperbolic GCNs.
}
, HAT \cite{zhang2019hyperbolic}.

\subsubsection{Parameter setting}
We perform a hyper-parameter search on a validation set for all methods.
The grid search is performed over the following search space:
Learning rate: [0.01, 0.008, 0.005, 0.001];
Dropout probability: [0.0, 0.1, 0.2, 0.3, 0.4, 0.5, 0.6, 0.7];
$L_2$ regularization strength: [0, 1e-1, 5e-2, 1e-2, 5e-3, 1e-3, 5e-4, 1e-4].
The results are reported over 10 random parameter initializations.
For all the methods, we set $d$, i.e., the dimension of latent representation as 8, 16, 32, 64 in link prediction and node classification tasks for a more comprehensive comparison.
In case studies, we set the dimension of latent representation as 64.
Note that the experimental setting in molecular property prediction task is same with \cite{liu2019hyperbolic}.
We optimize DeepWalk with SGD while optimize Poincar\'eEmb with RiemannianSGD \cite{bonnabel2013stochastic}.
The GCNs are optimized via Adam \cite{KingmaB14}.
Also,
LGCN leverages DropConnect \cite{wan2013regularization} which is the generalization of Dropout and can be used in the hyperbolic GCNs \cite{chami2019hyperbolic}.
Moreover,
although LGCN and the learned node representations are hyperbolic,
the trainable parameters in LGCN live in the tangent spaces,
which can be optimized via Euclidean optimization \cite{ganea2018hnn}, e.g., Adam \cite{KingmaB14}.
Furthermore,
LGCN uses early stopping based on validation set performance with a patience of 100 epochs.
The Hardware used in our experiments is:
Intel(R) Xeon(R) CPU E5-2620 v4 @ 2.10GHz,
GeForce @ GTX 1080Ti.

\begin{table*}
	\caption{Accuracy (\%) for node classification task. The best results are marked by bold numbers. }
	\centering
	\label{table_classify}
	\resizebox{0.89\textwidth}{!}{
		\begin{tabular}{cc| cc ccc ccc c}
			\toprule
			Dataset 	& dimension & deepwalk & poincar\'eEmb & GraphSage & GCN & GAT & HGCN & $\kappa$GCN & HAT & LGCN \\ \midrule        
			& 8  & 59.6±1.6 & 57.0±0.8 & 73.9±1.5 & 75.1±1.1 & 76.7±0.7 & 81.5±1.3 & 81.8±1.5 & 82.3±1.2 & \textbf{82.9±1.2} \\
			Disease 		& 16 & 61.5±2.2 & 56.1±0.7 & 75.3±1.0 & 78.3±1.0 & 76.6±0.8 & 82.8±0.8 & 82.1±1.1 & 83.6±0.9 & \textbf{84.4±0.8} \\
			$\delta_{avg}=0.00$ & 32 & 62.0±0.3 & 58.7±0.7 & 76.1±1.7 & 81.0±0.9 & 79.3±0.7 & 84.0±0.8 & 82.8±0.9 & 84.9±0.9 & \textbf{86.8±0.8} \\
			& 64 & 61.8±0.5 & 60.1±0.8 & 78.5±1.0 & 82.7±0.9 & 80.4±0.7 & 84.3±0.8 & 83.0±1.0 & 85.1±0.8 & \textbf{87.1±0.8} \\ \midrule
			& 8  & 44.3±0.6 & 38.9±1.1 & 46.8±1.4 & 50.5±0.5 & 47.8±0.7 & 50.5±1.1 & 49.1±0.9 & 50.7±1.0 & \textbf{51.6±1.1} \\
			USA 		& 16 & 42.3±1.3 & 38.3±1.0 & 47.5±0.8 & 50.9±0.6 & 49.5±0.7 & 51.1±1.0 & 50.5±1.2 & 51.3±0.9 & \textbf{51.9±0.9} \\
			$\delta_{avg}=0.16$ & 32 & 39.0±1.0 & 39.0±0.8 & 48.0±0.7 & 50.6±0.5 & 49.1±0.6 & 51.2±0.9 & 50.9±1.0 & 51.5±0.8 & \textbf{52.4±0.9} \\
			& 64 & 42.7±0.8 & 39.2±0.8 & 48.2±1.1 & 51.1±0.6 & 49.6±0.6 & 52.4±0.8 & 51.8±0.8 & 52.5±0.7 & \textbf{52.8±0.8} \\ \midrule
			& 8  & 66.7±1.0 & 65.3±1.1 & 71.3±1.6 & 70.9±1.1 & 70.0±0.9 & 71.7±1.3 & 70.3±1.2 & 71.0±1.0 & \textbf{72.0±1.3} \\ 
			Amazon 		& 16 & 67.5±0.8 & 67.0±0.7 & 72.3±1.6 & 70.9±1.1 & 72.7±0.8 & 72.7±1.3 & 71.9±1.1 & 73.3±1.0 & \textbf{75.0±1.1} \\ 
			$\delta_{avg}=0.20$ & 32 & 70.0±0.5 & 68.1±0.3 & 73.4±1.2 & 71.5±0.8 & 72.5±0.7 & 75.3±1.0 & 72.9±0.6 & 74.9±0.8 & \textbf{75.5±0.9} \\ 
			& 64 & 70.3±0.7 & 67.3±0.4 & 74.1±1.2 & 73.0±0.6 & 72.9±0.8 & 75.5±0.6 & 73.5±0.4 & 75.4±0.7 & \textbf{75.8±0.6} \\ \midrule
			& 8  & 64.5±1.2 & 57.5±0.6 & 74.5±1.3 & 80.3±0.8 & 80.4±0.8 & 80.0±0.7 & 81.0±0.5 & \textbf{82.8±0.7} & 82.6±0.8 \\ 
			Cora 		& 16 & 65.2±1.6 & 64.4±0.3 & 77.3±0.8 & 81.9±0.6 & 81.7±0.7 & 81.3±0.6 & 80.8±0.6 & 83.1±0.6 & \textbf{83.3±0.7} \\ 
			$\delta_{avg}=0.35$ & 32 & 65.9±1.5 & 64.9±0.4 & 78.8±1.2 & 81.5±0.4 & 82.6±0.7 & 81.7±0.7 & 81.8±0.5 & 83.2±0.6 & \textbf{83.5±0.6} \\ 
			& 64 & 66.5±1.7 & 68.6±0.4 & 79.2±0.6 & 81.6±0.4 & 83.1±0.6 & 82.1±0.7 & 81.5±0.7 & 83.1±0.5 & \textbf{83.5±0.5} \\ \midrule
			& 8  & 73.2±0.7 & 66.0±0.8 & 75.9±0.4 & 78.6±0.4 & 71.9±0.7 & 77.9±0.6 & 78.5±0.7 & 78.5±0.6 & \textbf{78.8±0.5} \\ 
			Pubmed 		& 16 & 73.9±0.8 & 68.0±0.4 & 77.3±0.3 & 79.1±0.5 & 75.9±0.7 & 78.4±0.4 & 78.3±0.6 & 78.6±0.5 & \textbf{78.6±0.7} \\ 
			$\delta_{avg}=0.36$ & 32 & 72.4±1.0 & 68.4±0.5 & 77.7±0.3 & 78.7±0.5 & 78.2±0.6 & 78.6±0.6 & 78.8±0.6 & 78.8±0.6 & \textbf{78.9±0.6} \\ 
			& 64 & 73.5±1.0 & 69.9±0.6 & 78.0±0.4 & 79.1±0.5 & 78.7±0.4 & 79.3±0.5 & 79.0±0.5 & 79.0±0.6 & \textbf{79.6±0.6} \\ \midrule
			& 8  & 47.8±1.6 & 38.6±0.4 & 65.8±1.6 & 68.9±0.7 & 69.5±0.8 & 70.9±0.6 & 70.3±0.6 & 71.2±0.7 & \textbf{71.8±0.7} \\ 
			Citeseer 	& 16 & 46.2±1.5 & 40.4±0.5 & 67.8±1.1 & 70.2±0.6 & 71.6±0.7 & 71.2±0.5 & 70.7±0.5 & \textbf{71.9±0.6} & \textbf{71.9±0.7} \\ 
			$\delta_{avg}=0.46$ & 32 & 43.6±1.9 & 43.5±0.5 & 68.5±1.3 & 70.4±0.5 & \textbf{72.6±0.7} & 71.9±0.4 & 71.2±0.5 & 72.4±0.5 & 72.5±0.5 \\ 
			& 64 & 46.6±1.4 & 43.6±0.4 & 69.2±0.8 & 70.8±0.4 & 72.4±0.7 & 71.7±0.5 & 71.0±0.3 & 72.2±0.5 & \textbf{72.5±0.6} \\ \bottomrule
		\end{tabular}
	}
\end{table*}

\subsection{Link prediction}
We compute the probability scores for edges by leveraging the Fermi-Dirac decoder \cite{krioukov2010hyperbolic,nickel2017poincare,chami2019hyperbolic}.
For the output node features $\mathbf{u}_i^{d,\beta}$ and $\mathbf{u}_j^{d,\beta}$,
the probability of existing the edge $e_{ij}$ between $\mathbf{u}_i^{d,\beta}$ and $\mathbf{u}_j^{d,\beta}$ is given as:
$p(\mathbf{u}_i^{d,\beta}, \mathbf{u}_i^{d,\beta})=
{1}/({e^{(d_\mathcal{L}^2(\mathbf{u}_i^{d,\beta}, \mathbf{u}_i^{d,\beta})-r)/t}+1}),$
where $r$ and $t$ are hyper-parameters.
We then minimize the cross-entropy loss to train the LGCN model.
Following \cite{chami2019hyperbolic}, 
the edges are split into 85\%, 5\%, 10\% randomly for training,
validation and test sets for all datasets,
and the evaluation metric is AUC.

The results are shown in Table \ref{table_linkpred}.
We can see that LGCN performs best in all cases, 
and its superiority is more significant for the low dimension setting.
Suggesting the graph operations of LGCN provide powerful ability to embed graphs.
Moreover,
hyperbolic GCNs perform better than Euclidean GCNs for datasets with lower $\delta_{avg}$,
which further confirms the capability of hyperbolic spaces in modeling tree-likeness graph data.
Furthermore,
compared with network embedding methods, 
GCNs achieve better performance in most cases,
which indicates GCNs can benefit from both structure and feature information in a graph.

\subsection{Node classification}

Here we evaluate the performance of LGCN on the node classification task.
We split nodes in Disease dataset into 30/10/60\% for training,
validation and test sets \cite{chami2019hyperbolic}.
For the other datasets, we use only 20 nodes per class for training, 
500 nodes for validation, 1000 nodes for test.
The settings are same with
\cite{chami2019hyperbolic,yang2016revisiting,kipf2016semi,velivckovic2017graph}.
The accuracy is used to evaluate the results.

Table \ref{table_classify} reports the performance.
We can observe similar results to Table \ref{table_linkpred}. 
That is,
LGCN preforms better than the baselines in most cases.
Also,
hyperbolic GCNs outperform Euclidean GCNs for datasets with lower $\delta_{avg}$,
and GCNs perform better than network embedding methods.
Moreover,
we notice that hyperbolic GCNs do not have an obvious advantage compared with Euclidean GCNs on Citeseer dataset,
which has the biggest $\delta_{avg}$. 
We think no obvious tree-likeness structure of Citeseer makes those hyperbolic GCNs do not work well on this task. 
In spite of this,
benefiting from the well-defined Lorentzian graph operations,
LGCN also achieves very competitive results.

\begin{table*}[!t]
	\hspace{-0.10\linewidth}
	\begin{minipage}[!t]{0.68\linewidth}
		
		\caption{The variants of LGCN and HGCN.}
		\centering
		\label{table_ablation}
		
		\resizebox{0.8\textwidth}{!}{
			\begin{tabular}{c cc cc cc}
				\toprule
				Dataset				& \multicolumn{2}{c}{Disease} & \multicolumn{2}{c}{USA} & \multicolumn{2}{c}{Amazon}\\
				Task				& LP	& NC	& LP	& NC	& LP	& NC	\\
				\midrule
				HGCN				& 92.7$\pm$0.4 & 84.3$\pm$0.8 &94.2$\pm$0.2 & 52.4$\pm$0.8 & 97.1$\pm$0.7 & 75.5$\pm$0.6\\
				HGCN$_{c}$			& 94.3$\pm$0.5 & 86.2$\pm$1.0 &95.6$\pm$0.1 & 52.5$\pm$0.8 & 97.3$\pm$0.3 & 75.8$\pm$0.4\\
				\midrule
				LGCN$_{\backslash\beta}$		& 96.3$\pm$0.4 & 86.3$\pm$0.7 & 96.1$\pm$0.3 & 52.5$\pm$0.7 & 96.5$\pm$0.7 & 75.6$\pm$0.5\\
				LGCN$_{\backslash att}$ 		& 95.9$\pm$0.3 & 86.6$\pm$0.8 & 95.9$\pm$0.2 & 52.2$\pm$0.7 & 97.0$\pm$0.6 & 74.6$\pm$0.5\\
				LGCN$_{\backslash att\backslash c}$	& 92.6$\pm$0.6 & 83.2$\pm$0.6 &	94.6$\pm$0.4 & 52.2$\pm$0.9	& 96.6$\pm$0.9 & 74.3$\pm$0.6\\
				LGCN				& 96.8$\pm$0.4 & 87.1$\pm$0.8 & 96.4$\pm$0.2 & 52.8$\pm$0.8 & 97.6$\pm$0.5 & 75.8$\pm$0.6\\
				\bottomrule
			\end{tabular}
		}
	\end{minipage}
	\begin{minipage}[!t]{0.32\linewidth}
		\caption{MAE (scaled by 100) of predicting molecular properties logP, QED and SAS on ZINC dataset.}
		\centering
		\label{table_graph_regression}
		\resizebox{1.\textwidth}{!}{
			\begin{tabular}{cc cc}
				\toprule
				Property & logP 	& QED 	& SAS\\
				\midrule
				MPNN\cite{gilmer2017neural}		& 4.1±0.02 & 8.4±0.05 & 9.2±0.07\\
				GGNN\cite{li2015gated}		& 3.2±0.20 & 6.4±0.20 & 9.1±0.10\\
				HGNN$_P$	& 3.1±0.01 & 6.0±0.04 & 8.6±0.02\\
				HGNN$_H$	& 2.4±0.02 & 4.7±0.00 & 7.7±0.06\\
				HGNN$_{L}$		& \textbf{2.2±0.03} & \textbf{3.3±0.05} & \textbf{5.8±0.05}\\
				\bottomrule
			\end{tabular}
		}
	\end{minipage}
\end{table*}

\subsection{Analysis}

\subsubsection{Ablations study}
Here we evaluate the effectiveness of some components in LGCN,
including self attention ($att$) and trainable curvature ($\beta$).
We remove these two components from LGCN and obtain two variants LGCN$_{\backslash att}$ and LGCN$_{\backslash\beta}$, respectively.
To further validate the performance of the proposed centroid-based Lorentzian aggregation,
we exchange the aggregation of HGCN and LGCN$_{\backslash att}$,
denoted as  HGCN$_c$ and LGCN$_{\backslash att\backslash c}$, respectively.
To better analyze the ability of modeling graph with underlying hyperbolic geometry,
we conduct the link prediction (LP) and node classification (NC) tasks on three datasets with lower $\delta_{avg}$, i.e., Disease, USA, and Amazon datasets.
The results are shown in Table \ref{table_ablation}.
Comparing LGCN to its variants,
we observe 
that LGCN 
always achieves best performances, indicating the effectiveness of self attention and trainable curvature. 
Moreover,
HGCN$_{c}$ achieves better results than HGCN, while LGCN$_{\backslash att}$ performs better than LGCN$_{\backslash att\backslash c}$ in most cases,
suggesting the effectiveness of the proposed neighborhood aggregation method.


\begin{figure*} 
	\hspace{-0.03\linewidth}
	\begin{minipage}[t]{0.22\linewidth} 
		\centering 
		\includegraphics[width=1.1in]{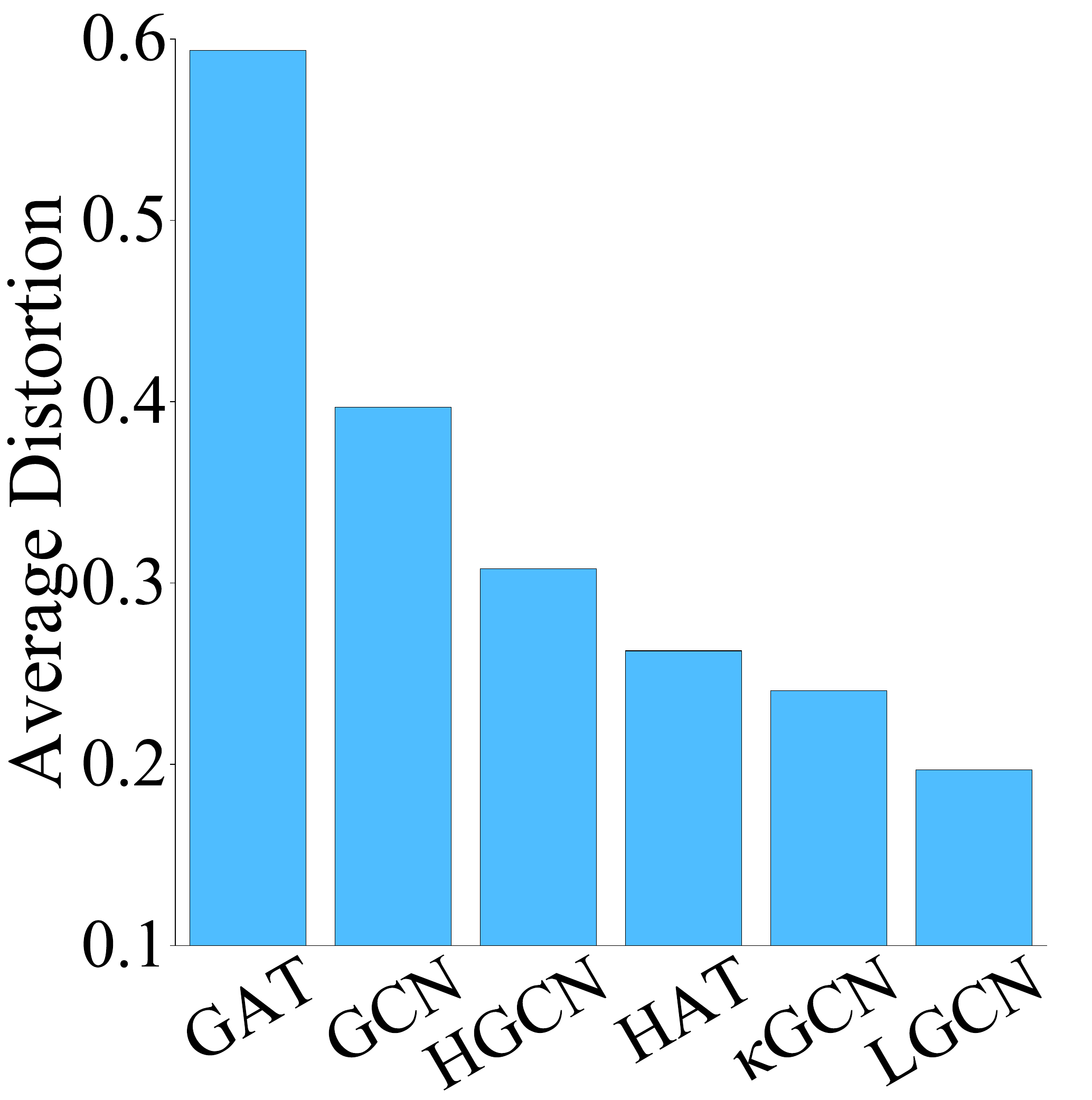} 
		\caption{Distortion of  $\qquad$ different methods on $\qquad$ Disease.} 
		\label{fig_distortion} 
	\end{minipage}%
	\begin{minipage}[t]{0.24\linewidth} 
		\centering 
		\includegraphics[width=1.1in]{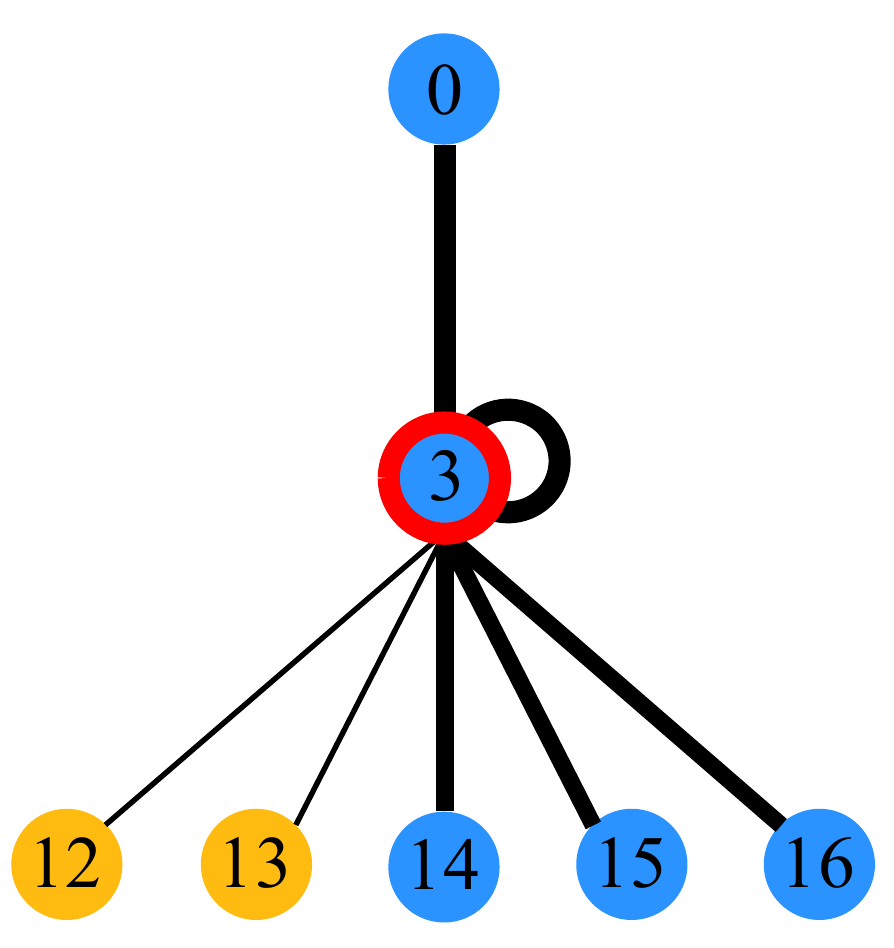} 
		\caption{Attention value  $\qquad$ on Disease.} 
		\label{fig_att} 
	\end{minipage}%
	\begin{minipage}[t]{0.24\linewidth} 
		\centering 
		\includegraphics[width=1.42in]{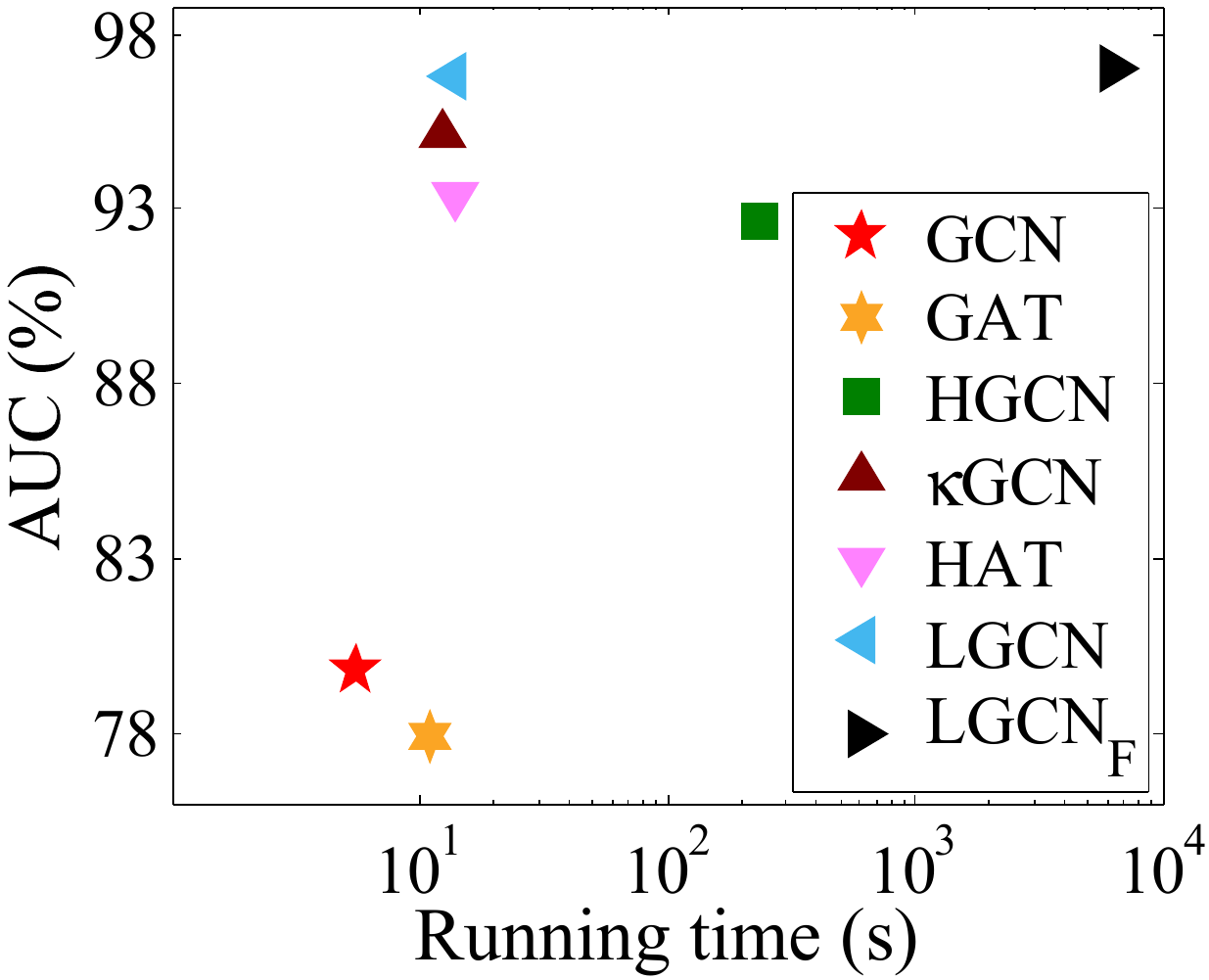} 
		\caption{Performance v.s. \\ the running time on Disease.}
		\label{fig_time} 
	\end{minipage} 
	\begin{minipage}[t]{0.24\linewidth} 
		\centering 
		\includegraphics[width=1.45in]{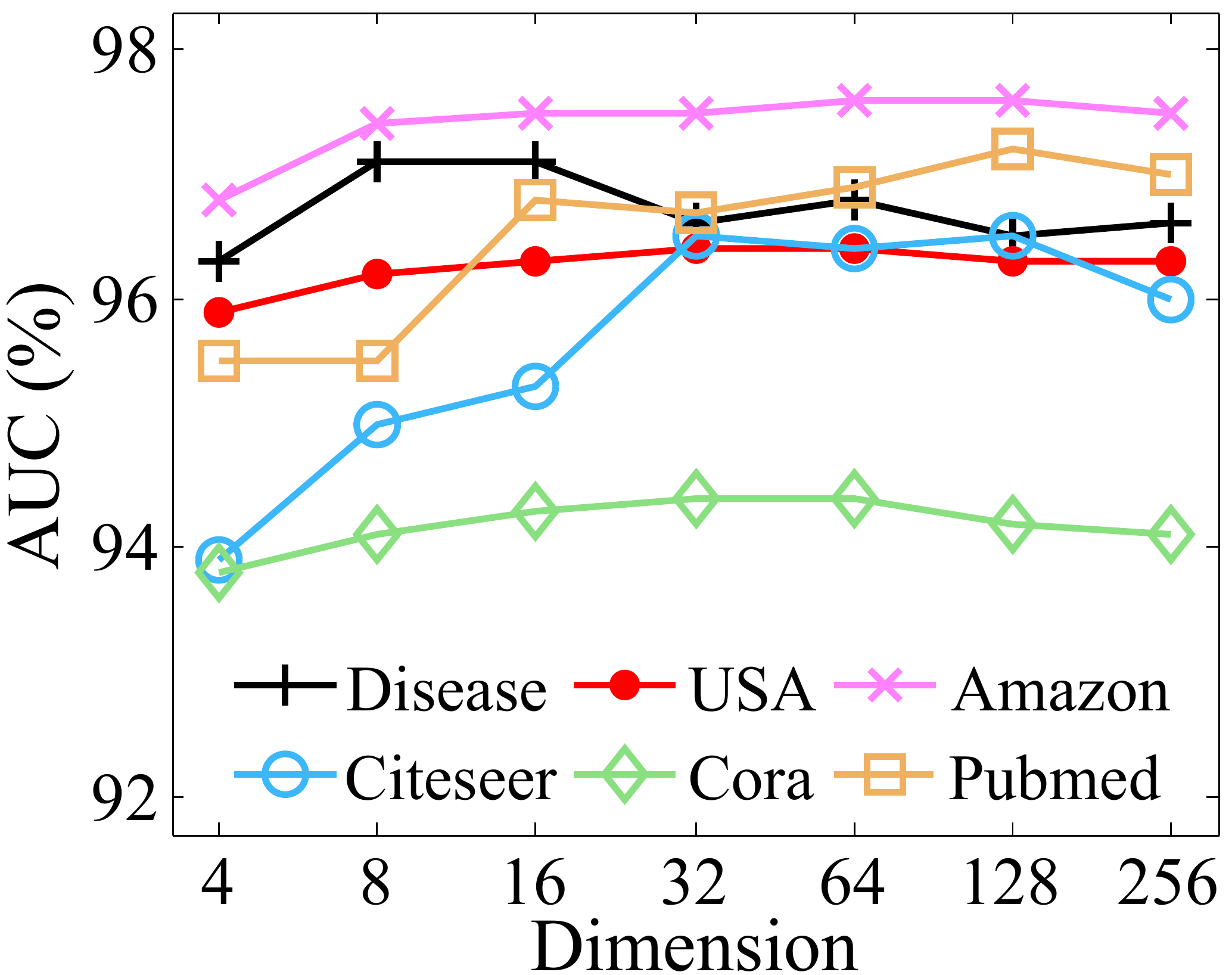} 
		\caption{Results w.r.t. attention matrix dimension  on different datasets.}
		\label{fig_dim} 
	\end{minipage} 
\end{figure*}

\subsubsection{Extension to graph-level task: molecular property prediction}\label{sec_graph_regression}
Most existing hyperbolic GCNs focus on node-level tasks, 
e.g., node classification \cite{chami2019hyperbolic,bachmann2019constant,zhang2019hyperbolic} and link prediction \cite{chami2019hyperbolic} tasks.
We also notice that HGNN \cite{liu2019hyperbolic}, as a hyperbolic GCN,
achieves good results on graph-level task, i.e., molecular property prediction.
Here we provide a Lorentzian version of HGNN named HGNN$_L$,
which keeps the model structure of HGNN and replaces its graph operations with those operations defined in this paper,
i.e., feature transformation, neighborhood aggregation, and non-linear activation.
Following HGNN \cite{liu2019hyperbolic},
we conduct molecular property prediction task on ZINC dataset.
The experiment 
is a regression task to predict molecular properties:
the water-octanal partition coefficient (logP), 
qualitative estimate of drug-likeness (QED), 
and synthetic accessibility score (SAS).
The experimental setting is same with HGNN for a fair comparison,
and we reuse the metrics already reported in HGNN for state-of-the-art techniques.
HGNN implemented with Poincar\'e ball, hyperboloid model is denoted as HGNN$_P$, HGNN$_H$, respectively.
The results of mean absolute error (MAE) are shown in Table \ref{table_graph_regression}.
HGNN$_L$ achieves best performance among all the baselines.
Also, HGNN$_L$ can further improve the performance of HGNN,
which verifies the effectiveness of the proposed graph operations.
Moreover,
as HGNN$_L$ is obtained by simply replacing the graph operation of HGNN$_H$,
the proposed Lorentzian operations provide an alternative way for hyperbolic deep learning.


\subsubsection{Distortion analysis}
We have mentioned that some existing hyperbolic GCNs could cause a distortion to the learned graph structure,
since their graph operations do not rigorously follow the hyperbolic geometry.
Thus, here we evaluate the distortions of GCNs on the Disease dataset.
Following \cite{gu2018learning},
we define the average distortion to evaluate the ability of GCNs to preserve graph structures.
The average distortion is defined as:
$
\frac{1}{n^2} \sum_{i,j} \Big( \big( \frac{d(\mathbf{u}_i, \mathbf{u}_j)/ d_{avg} }{d^G(i, j)/d^G_{avg}} \big)^2 - 1 \Big)^2,
$
where $d(\mathbf{u}_i, \mathbf{u}_j)$ is the intrinsic distance 
between the output features of nodes $i$ and $j$, 
and $d^G(i,j)$ is their graph distance. 
Both of the distances are divided by their average values, i.e., $d_{avg}$ and $d^G_{avg}$, to satisfy the scale invariance.
The results of the link prediction are shown in Fig. \ref{fig_distortion},
and a lower average distortion indicates a better preservation of the graph structure.
We can find that LGCN has the lowest average distortion among these GCNs,
which is benefited from the well-defined graph operations.
Also, all the hyperbolic GCNs have lower average distortion compared with Euclidean GCNs.
That is also reasonable, since hyperbolic spaces is more suitable to embed tree-likeness graph than Euclidean spaces.


\subsubsection{Attention analysis}
In addition, we examine the learned attention values in LGCN.
Intuitively,
important neighbors tend to have large attention values.
We take a node in the Disease dataset for node classification task as an illustrative example.
As shown in Fig. \ref{fig_att},
the nodes are marked by their indexes in the dataset,
and the node with a red outline is the center node.
The color of a node indicates its label,
and the attention value for a node is visualized by its edge width.
We observe that the center node $3$ pays more attention to nodes with the same class, i.e., nodes $0, 3, 14, 15, 16$,
suggesting that the proposed attention mechanism can automatically distinguish 
the difference among neighbors and assign the higher weights
to the meaningful neighbors.


\subsubsection{Efficiency comparison}
We further analyze the efficiency of some GCNs.
To better analyze the aggregation in Theorem \ref{thm_agg},
we provide a variant of LGCN named LGCN$_{F}$,
which minimizes Eq. \eqref{eq_except_agg} w.r.t. the intrinsic distance, i.e., Eq. \eqref{eq:intrinsic_distance}.
Note that the aggregation of LGCN$_{F}$ is a kind of Fr\'echet mean which dose not have closed-form solutions,
so we compute it via a state-of-the-art gradient descent based method \cite{lou2020differentiating}.
Here we report the link prediction performance and training time per 100 epochs of GCNs on Disease dataset in Fig. \ref{fig_time}.
One can see that GCN is the fastest.
Most hyperbolic GCNs, e.g., $\kappa$GCN, HAT, LGCN,
are on the same level with GAT.
HGCN is slower than above methods,
and LGCN$_F$ is the slowest.
Although hyperbolic GCNs are slower than Euclidean GCNs,
they have better performance.
Moreover,
HGCN is significantly slower than LGCN,
since HGCN aggregates different nodes in different tangent spaces, and this process cannot be computed parallelly.
HAT addresses this problem by aggregating all the nodes in the same tangent space.
Despite the running time of HAT and $\kappa$GCN are on par with LGCN,
LGCN achieves better results.
Furthermore,
both LGCN and LGCN$_F$ aggregate nodes by minimizing an expectation of distance.
However, the aggregation in LGCN has a closed-form solution while LGCN$_F$ has not.
Despite LGCN$_F$ has a little improvement with LGCN,
it is not cost-effective.
To sum up,
LGCN can learn more effective node representations with acceptable efficiency.

\subsubsection{Parameter sensitivity}
We further test the impact of attention matrix size of LGCN.
We change the horizontal dimension of matrix from 4 to 256.
The results on the link prediction task are shown in Fig. \ref{fig_dim}.
We can see that with the growth of the matrix size, the performance raises first and then starts to drop slowly.
The results indicate that the attention matrix needs a suitable size to learn the attention coefficient.
Also, LGCN has a stable perofmance when the horizontal dimension ranges from 32 to 128.


\section{Conclusion}
Existing hyperbolic GCNs cannot rigorously follow the hyperbolic geometry, which might limit the ability of hyperbolic geometry.
To address this issue,
we propose a novel Lorentzian graph neural network, called LGCN,
which designs rigorous hyperbolic graph operations, e.g., feature transformation and non-linear activation.
An elegant neighborhood aggregation method is also leveraged in LGCN,
which conforms to the mathematical meanings of hyperbolic geometry.
The extensive experiments
demonstrate the superiority of LGCN, compared with the state-of-the-art methods.

\begin{acks}
	This work is supported by the National Natural Science Foundation of China (No. U20B2045, 61772082, 62002029, U1936104, 61972442), and the National Key Research and Development Program of China (2018YFB1402600). It is also supported by BUPT Excellent Ph.D. Students Foundation (No. CX2019126).
\end{acks}

\bibliographystyle{ACM-Reference-Format}
\bibliography{sample-base}

%
%
%
%
%
%
%
%
\appendix

\section{Hyperbolic geometry}
Hyperbolic geometry is a non-Euclidean geometry with a constant negative curvature.
There are some equivalent hyperbolic models to describe hyperbolic geometry, 
including the hyperboloid model, the Poincar\'e ball model, and the Klein ball model, etc.
Since we have introduced the hyperboloid model in Section \ref{sec:hyperbolic_geometry},
here we introduce the Poincar\'e ball model.
More rigorous and in-depth introduction of differential geometry and hyperbolic geometry can be found in \cite{helgason1979differential,lee2009manifolds,ratcliffe1994foundations}.

\textbf{Poincar\'e ball.} 
We consider a specific Poincar\'e ball model $\mathbb{D}^{n, \alpha}$ \cite{ganea2018hnn}, which is defined by an open $d$-dimensional ball of radius $1/\sqrt{\alpha}$ ($\alpha>0$):
$
\mathbb{D}^{n,\alpha}:=\{\mathbf{x}\in \mathbb{R}^n:\alpha\|\mathbf{x}\|^2<1 \},
$
equipped with the  Riemannian metric: $g^{\mathbb{D}}_\mathbf{x}=({\lambda^\alpha_\mathbf{x}})^2g^\mathbb{R}$, where $\lambda_{\mathbf{x}}^\alpha = 2/(1-\alpha\|\mathbf{x}\|^2)$, $g^\mathbb{R}=\mathbf{I}_d$.
When $\mathbf{x}=\mathbf{0}\in\mathbb{D}^{n,\alpha}$,
the exponential map $\exp_\mathbf{0}^\alpha: \mathcal{T}_\mathbf{0}\mathbb{D}^{n,\alpha}\rightarrow\mathbb{D}^{n,\alpha}$ and the logarithmic map
$\log_\mathbf{0}^\alpha: \mathbb{D}^{n,\alpha} \rightarrow \mathcal{T}_\mathbf{0}\mathbb{D}^{n,\alpha}$ 
are given for 
$\mathbf{v}\!\!\in\!\!\mathcal{T}_\mathbf{0}\mathbb{D}^{n,\alpha}\!\backslash\!\{\mathbf{0}\}$ and $\mathbf{y}\in\mathbb{D}^{n,\alpha}\backslash\{\mathbf{0}\}$:
\begin{flalign}
	\exp_{\mathbf{0}}^\alpha(\mathbf{v})&=\tanh(\sqrt{\alpha}\|\mathbf{v}\|)\frac{\mathbf{v}}{\sqrt{\alpha}\|\mathbf{v}\|},\\
	\log_{\mathbf{0}}^\alpha\!({\mathbf{y}}) &= \tanh^{-1}\!(\sqrt{\alpha}\|{\mathbf{y}}\|)\frac{{\mathbf{y}}}{\sqrt{\alpha}\|{\mathbf{y}}\|}.
\end{flalign}
The Poincar\'e ball model and the hyperboloid model are isomorphic, and the diffeomorphism maps one space onto the other as shown in following \cite{chami2019hyperbolic}:
\begin{flalign}
	p_{\mathbb{H}^{n,\beta}\rightarrow \mathbb{D}^{n,\alpha}}(x_0, x_1, \cdots, x_n)&=
	\frac{\sqrt{\beta}(x_1, \cdots, x_n)}
	{\sqrt{\beta} + x_o},\label{eq_hyper2poincare}
	\\
	p_{\mathbb{D}^{n,\alpha}\rightarrow\mathbb{H}^{n,\beta} }(x_1, \cdots, x_n)&=
	\frac{(1/\sqrt{\alpha}+\sqrt{\alpha}\|\mathbf{x}\|_2^2, 2x_1, \cdots, 2x_n)}
	{1-\alpha\|\mathbf{x}\|^2}.\label{eq_poincare2hyper}
\end{flalign}

\section{Proofs of results}

\subsection{Proof of Theorem 3.1}\label{sec:app_proof3.1}

\textit{Proof}.
Let $\mathbf{x}^{n,\beta}\in\mathbb{H}^{n,\beta}$, $\mathbf{v}=(v_0,v_1,\cdots,v_n)\in \mathcal{T}_\mathbf{0}\mathbb{H}^{n,\beta}$, and $\mathbf{M}$ be a $m\times n$ matrix,
Lorentzian matrix-vector multiplication is shown as following:
\begin{equation}
	\begin{split}
		\mathbf{M}{\otimes^\beta}\mathbf{x}^{n,\beta}:&=\mathbf{M}^{\otimes^\beta}(\mathbf{x}^{n,\beta})=\exp_\mathbf{0}^\beta\Big(\hat{\mathbf{M}}\big(\log_\mathbf{0}^\beta(\mathbf{x}^{n,\beta})\big)\Big)=\mathbf{y}^{m,\beta},\\
		\hat{\mathbf{M}}(\mathbf{v})&=\big(0, \mathbf{M}(v_1,\cdots,v_n)\big),
	\end{split}
\end{equation}
Let $\mathbf{x}^{n,\alpha}\in\mathbb{D}^{n,\beta}$,
M\"obius matrix-vector multiplication has the formulation as \cite{ganea2018hnn}:
\begin{equation}
	\begin{split}
		&\mathbf{M}\otimes^{\alpha}\mathbf{x}^{n,\alpha}:\\
		=&(1/\sqrt{\alpha})\tanh\Big( \frac{\|\mathbf{M}\mathbf{x}^{n,\alpha}\|}{\|\mathbf{x}^{n,\alpha} \|} \tanh^{-1}(\sqrt{\alpha}\|\mathbf{x}^{n,\alpha}\|) \Big)\frac{\mathbf{Mx}^{n,\alpha}}{\|\mathbf{Mx}^{n,\alpha}\|}
		=\mathbf{y}^{m,\alpha}.
	\end{split}
\end{equation}
For $p_{\mathbb{H}^{n,\beta}\rightarrow\mathbb{D}^{n,\alpha}}(\mathbf{x}^{n,\beta})=\mathbf{x}^{n,\alpha}$ and a shared $m\times n$ matrix $\mathbf{M}$, we aim to prove $p_{\mathbb{H}^{n,\beta}\rightarrow\mathbb{D}^{n,\alpha}}(\mathbf{y}^{n,\beta})=\mathbf{y}^{n,\alpha}.$

For $\mathbf{x}^{n,\beta}\!=\!(x_0^{(\beta)}\!, x_1^{(\beta)}\!,\cdots\!,x_n^{(\beta)})\in\mathbb{H}^{n,\beta}$,
let
$\hat{\mathbf{x}}^{(\beta)}\!=\!(x_1^{(\beta)}\!,\cdots\!,x_n^{(\beta)})$,
and
the logarithmic map of $\mathbf{x}$ at $\mathbf{0}^{n,\beta}=(\sqrt{\beta}, 0, \cdots, 0)\in\mathbb{H}^{n,\beta}$, i.e.,  is shown as following:
\begin{equation}\label{eq_prove_log_x}
	\begin{split}
		&\log_{\mathbf{0}}^{\beta}(\mathbf{x}^{n,\beta})
		=\sqrt{\beta}\cosh^{-1}\big(\frac{x_0^{(\beta)}}{\sqrt{\beta}}\big)\frac{(0,\hat{\mathbf{x}}^{(\beta)})}{\|\hat{\mathbf{x}}^{(\beta)}\|}.
	\end{split}
\end{equation}
Let $q=\sqrt{\beta}\cosh^{-1}(x_0^{\beta}/\sqrt{\beta})/\|\hat{\mathbf{x}}^{(\beta)}\|$,
$\log_{\mathbf{0}}^{\beta}(\mathbf{x}^{n,\beta})=q(0,\hat{\mathbf{x}}^{(\beta)})$,
so we have:
\begin{equation}
	\hat{\mathbf{M}}\big(\log_{\mathbf{0}}^\beta(\mathbf{x}^{n,\beta})\big)= (0, q\mathbf{M\hat{x}}^{(\beta)}) = \mathbf{m}.
\end{equation}
The Lorentzian matrix-vector multiplication is given as following:
\begin{equation}
	\begin{split}
		\mathbf{M}{\otimes^\beta}\mathbf{x}^{n,\beta}
		&\!\!=\!\cosh(\frac{\|\mathbf{m}\|_\mathcal{L}}{\sqrt{\beta}})\cdot\mathbf{0}^{n,\beta}+\sqrt{\beta}\sinh(\frac{\|\mathbf{m}\|_\mathcal{L}}{\sqrt{\beta}})\frac{\mathbf{m}}{\|\mathbf{m}\|_\mathcal{L}}\\
		&\!\!=\!\Big( \sqrt{\beta} \cosh\big(\frac{\|\mathbf{m}\|_\mathcal{L}}{\sqrt{\beta}}\big),\frac{\sqrt{\beta}\sinh\big(\frac{\|\mathbf{m}\|_\mathcal{L}}{\sqrt{\beta}}\big)q}{\|\mathbf{m}\|_\mathcal{L}} \!\cdot\! \mathbf{M\hat{x}}^{(\beta)} \!\Big)
		\!=\!\mathbf{y}^{m,\beta}.
	\end{split}
\end{equation}
Then we map $\mathbf{y}^{m,\beta}$ to the Poincar\'e ball via Eq. \eqref{eq_hyper2poincare},
\begin{equation}\label{eq_p2h_y}
	\begin{split}
		&p_{\mathbb{H}^{n,\beta}\rightarrow \mathbb{D}^{n,\alpha}}(\mathbf{y}^{m,\beta})
		=\frac{\sinh\big(\frac{\|\mathbf{m}\|_\mathcal{L}}{\sqrt{\beta}}\big)}{1+\cosh\big(\frac{\|\mathbf{m}\|_\mathcal{L}}{\sqrt{\beta}}\big)}\cdot\frac{\sqrt{\beta}}{\|\mathbf{M\hat{x}}^{(\beta)}\|}\cdot\mathbf{M\hat{x}}^{(\beta)}\\
		=&\tanh\Big( \frac{\|\mathbf{M\hat{\mathbf{x}}^{(\beta)}}\|}{2\|\hat{\mathbf{x}}^{(\beta)}\|}\cdot\cosh^{-1}\big( \sqrt{\frac{\beta+\|\hat{\mathbf{x}}^{(\beta)}\|^2}{{\beta}}} \big) \Big)\cdot\frac{\sqrt{\beta}}{\|\mathbf{M\hat{x}}^{(\beta)}\|}\cdot\mathbf{M\hat{x}}^{(\beta)}.
	\end{split}
\end{equation}
Note that $\|\mathbf{m}\|_\mathcal{L}\!\!=\!\!\sqrt{\langle\mathbf{m},\mathbf{m}\rangle_\mathcal{L}}\!\!=\!\!\|q\mathbf{M\hat{x}}^{(\beta)}\|$
and $x_0^{(\beta)}\!\!=\!\!\sqrt{{\beta}+\|\hat{\mathbf{x}}^{(\beta)}\|^2}.$
Moreover, the point
$\mathbf{x}^{n,\alpha}=( x_1^{(\alpha)},\cdots,x_n^{(\alpha)})\in\mathbb{D}^{n,\alpha}$ can be mapped into the hyperboloid model via Eq. \eqref{eq_poincare2hyper} as following:
\begin{equation}\label{eq_poincare2lorentz}
	\begin{split}
		p_{\mathbb{D}^{n,\alpha}\rightarrow \mathbb{H}^{n,\beta}}(\mathbf{x}^{n,\alpha})&=\frac{(1/\sqrt{\alpha}+\sqrt{\alpha}\|\mathbf{x}^{n,\alpha}\|_2^2, 2x_1^{(\alpha)}, \cdots, 2x_n^{(\alpha)})}
		{1-\alpha\|\mathbf{x}^{n,\alpha}\|^2}\\
		&=(x_0^{(\beta)}, x_1^{(\beta)},\cdots,x_n^{(\beta)}).
	\end{split}
\end{equation}
Thus, the squared norm of $\hat{\mathbf{x}}^{(\beta)}$ is given as:
\begin{equation}\label{eq_x_norm}
	\begin{split}
		\|\hat{\mathbf{x}}^{(\beta)}\|^2
		=\sum_{i=1}^n (x_i^{(\beta)})^2
		&=\Big( \frac{2\|\mathbf{x}^{n,\alpha}\|}{1-\alpha\|\mathbf{x}^{n,\alpha}\|^2} \Big)^2.
	\end{split}
\end{equation} 
Moreover, the curvature of the Poincar\'e ball model $\mathbb{D}^{n,\alpha}$ is $-\alpha$,
while the curvature of the hyperboloid model $\mathbb{H}^{n,\beta}$ is $-\beta$.
The maps between the two models in Eq. \eqref{eq_hyper2poincare} and Eq. \eqref{eq_poincare2hyper} ensure they have a same curvature, i.e., $-\alpha=-1/\beta$, $\alpha=1/\beta$.
Therefore, combining Eq. \eqref{eq_p2h_y} and Eq. \eqref{eq_x_norm}, we have:
\begin{equation}
	\begin{split}
		&p_{\mathbb{H}^{m,\beta}\rightarrow \mathbb{D}^{m,\alpha}}(\mathbf{y}^{m,\beta})\\
		=&\frac{1}{\sqrt{\alpha}}\tanh\left( \frac{\|\mathbf{M{\mathbf{x}}}^{n,\alpha}\|}{2\|\mathbf{x}^{n,\alpha}\|}\cdot \cosh^{-1}{\Big(\frac{1+\alpha\|\mathbf{x}^{n,\alpha}\|^2}{1-\alpha\|\mathbf{x}^{n,\alpha}\|^2}}\Big) \right)\cdot\frac{\mathbf{Mx}^{n,\alpha}}{\|\mathbf{Mx}^{n,\alpha}\|}\\
		=&\frac{1}{\sqrt{\alpha}}\!\tanh\!\left(\! \frac{\|\mathbf{M{\mathbf{x}}}^{n,\alpha}\|}{\|\mathbf{x}^{n,\alpha}\|}\!\cdot\! \tanh^{-1}(\sqrt{\alpha}\|\mathbf{x}^{n,\alpha}\|)  \!\right)\!\cdot\!\frac{\mathbf{Mx}^{n,\alpha}}{\|\mathbf{Mx}^{n,\alpha}\|}
		\!=\!\mathbf{y}^{m,\alpha}.
	\end{split}
\end{equation}
Therefore, Lorentzian matrix-vector multiplication is equivalence to M\"obius matrix-vector multiplication.$\hfill\square$

\subsection{Proof of Theorem 3.2}\label{sec:app_proof3.2}

\textit{Proof}.
Combining Eq. \eqref{eq_except_agg} and Eq. \eqref{eq_lorentz_dis},
we have following equality:
\begin{equation}
	\begin{split}
		&\arg\min _{\mathbf{c}^{d,\beta}\in\mathbb{H}^{d,\beta}}\sum_{j\in N(i) \cup \{i\}} w_{ij}d_{\mathcal{L}}^2(\mathbf{h}_j^{d,\beta}, \mathbf{c}^{d,\beta})\\
		=&\arg\max_{\mathbf{c}^{d,\beta}\in\mathbb{H}^{d,\beta}}\sum_{j\in N(i) \cup \{i\}} w_{ij}\langle\mathbf{h}_j^{d,\beta}, \mathbf{c}^{d,\beta}\rangle_\mathcal{L}\\
		=&\arg\max_{\mathbf{c}^{d,\beta}\in\mathbb{H}^{d,\beta}} \langle\mu\sum_{j\in N(i) \cup \{i\}} w_{ij}\mathbf{h}_j^{d,\beta}, \mathbf{c}^{d,\beta}\rangle_\mathcal{L},
	\end{split}
\end{equation}
where $\mu>0$ is a scaling factor to satisfy $\mu\sum_{j\in N(i) \cup \{i\}} w_{ij}\mathbf{h}_j^{d,\beta}\in\mathbb{H}^{n,\beta}$.
Also, we can infer from Eq. \eqref{eq_l_inner_eq},
$\langle\mu\sum_{j\in N(i) \cup \{i\}}\!\! w_{ij}\mathbf{h}_j^{d,\beta}\!\!,\! \mathbf{c}^{d,\beta}\rangle_\mathcal{L}$ $\leq-\beta$ 
and
$\langle\mu\!\sum_{j\in N(\!i\!) \cup \{\!i\!\}} \!\!w_{ij}\mathbf{h}_j^{d,\beta}\!,\! \mathbf{c}^{d,\beta}\rangle_\mathcal{\!\!L}\!\!=\!\!-\beta$ 
iff 
$\mu\!\!\sum_{j\in N(\!i\!) \cup \{\!i\!\}}\! w_{ij}\mathbf{h}_j^{d,\beta}\!\!\!=\mathbf{c}^{d,\beta}$,
we need to find a $\mu$ to satisfy $\mu\sum_{j\in N(i) \cup \{i\}} w_{ij}\mathbf{h}_j^{d,\beta}=\mathbf{c}^{d,\beta}$.
Assuming that $\mu_0>0$ satisfies $\mu_0\sum_{j\in N(i) \cup \{i\}} w_{ij}\mathbf{h}_j^{d,\beta}=\mathbf{c}^{d,\beta}$,
so $\mu_0\sum_{j\in N(i) \cup \{i\}} w_{ij}\mathbf{h}_j^{d,\beta}\in\mathbb{H}^{n,\beta}$.
Thus the Lorentzian scalar product of it and itself should equal to $-\beta$,
and we have:
\begin{equation}
	\begin{split}
		\mu_0^2\langle \sum_{j\in N(i) \cup \{i\}} w_{ij}\mathbf{h}_j^{d,\beta}, \sum_{j\in N(i) \cup \{i\}} w_{ij}\mathbf{h}_j^{d,\beta} \rangle_\mathcal{L}&=-\beta\\
		|\mu_0^2 \|\sum_{j\in N(i) \cup \{i\}} w_{ij}\mathbf{h}_j^{d,\beta}\|^2_\mathcal{L}|&=|-\beta|\\
		\mu_0^2 |\|\sum_{j\in N(i) \cup \{i\}} w_{ij}\mathbf{h}_j^{d,\beta}\|^2_\mathcal{L}|&=\beta.
	\end{split}
\end{equation}
We have 
$\mu_0^2\!=\!\!\frac{\beta}{|\|\sum_{j\in N(i) \cup \{i\}} \!\!w_{ij}\mathbf{h}_j^{d,\beta}\|^2_\mathcal{\!L}\!|}$, 
and
$\mu_0\!\!=\!\!\frac{\sqrt{\beta}}{|\|\sum_{j\in N(i) \cup \{i\}} \!\!w_{ij}\mathbf{h}_j^{d,\beta}\|_\mathcal{\!L}\!|}\!\!>0$.
Therefore, we have result:
$
\mathbf{c}^{d,\beta}=\sqrt{\beta}\frac{\sum_{{j\in N(i) \cup \{i\}}}w_{ij}\mathbf{h}_j^{d,\beta}}{|\|\sum_{{j\in N(i) \cup \{i\}}}w_{ij}\mathbf{h}_j^{d,\beta}\|_{\mathcal{L}}|}.
$
Moreover,
since for any $j$, $\langle\mathbf{h}_j^{d,\beta},\mathbf{h}_j^{d,\beta}\rangle_\mathcal{L}=-\beta$,
it satisfies that
$\langle\sum_{{j\in N(i) \cup \{i\}}}w_{ij}\mathbf{h}_j^{d,\beta},\sum_{{j\in N(i) \cup \{i\}}}w_{ij}\mathbf{h}_j^{d,\beta} \rangle_\mathcal{L}<0,$
so it is easy to check that
$\langle\! \sqrt{\beta}
\frac{\sum_{{j\in N(i) \cup 
			\{i\}}}\!\!w_{ij}\mathbf{h}_j^{d,\beta}}{|\|\sum_{{j\in N(i) \cup \{i\}}}\!\!w_{ij}\mathbf{h}_j^{d,\beta}\|_{\mathcal{\!L}}|},\sqrt{\beta}\frac{\sum_{{j\in N(i) \cup \{i\}}}\!\!w_{ij}\mathbf{h}_j^{d,\beta}}{|\|\sum_{{j\in N(i) \cup \{i\}}}\!\!w_{ij}\mathbf{h}_j^{d,\beta}\|_{\mathcal{\!L}}|} \rangle_\mathcal{\!\!L}\!\!=-\beta$,
and $\mathbf{c}^{d,\beta}\in\mathbb{H}^{n,\beta}$.
$\hfill\square$ 

\subsection{Proof of Theorem 3.3}\label{sec:app_proof3.3}

\textit{Proof}.
Here we prove the theorem in the case of leveraging LeaklyRelu activation function, as an example.
Let $\mathbf{x}^{n,\beta}\in\mathbb{H}^{n,\beta}$, $\mathbf{v}=(v_0,v_1,\cdots,v_n)\in \mathcal{T}_\mathbf{0}\mathbb{H}^{n,\beta}$, 
$\sigma(\cdot)$ be the LeaklyRelu activation function,
and Lorentzian non-linear activation is given as following:
\begin{flalign}
	\sigma^{\otimes^\beta}(\mathbf{x}^{n,\beta})&=\exp_\mathbf{0}^\beta\Big(\hat{\sigma}^{\otimes^\beta}\big(\log_\mathbf{0}^\beta(\mathbf{x}^{n,\beta})\big)\Big)=\mathbf{y}^{n,\beta},\\
	\hat{\sigma}^{\otimes^\beta}(\mathbf{v})&=\big(0, \sigma(v_1),\cdots,\sigma(v_n)\big),\label{eq_prove_l_non2}
\end{flalign}
where $\sigma(v_i)=\max(k v_i, v_i)$ for $k\in(0,1)$.
Let $\mathbf{x}^{n,\alpha}\in\mathbb{D}^{n,\beta}$,
M\"obius non-linear activation has the formulation as \cite{ganea2018hnn}:
\begin{equation}\label{eq_pr_m_nonl}
	\sigma^{\otimes^\alpha}(\mathbf{x}^{n,\alpha})=\exp_{\mathbf{0}}^\alpha\Big( \sigma\big(\log_{\mathbf{0}}^\alpha(\mathbf{x}^{n,\alpha})\big) \Big)=\mathbf{y}^{n,\alpha}.
\end{equation}
For $p_{\mathbb{H}^{n,\beta}\rightarrow\mathbb{D}^{n,\alpha}}(\mathbf{x}^{n,\beta})=\mathbf{x}^{n,\alpha}$ and the LeaklyRelu activation function $\sigma(\cdot)$, 
we aim to prove $p_{\mathbb{H}^{n,\beta}\rightarrow\mathbb{D}^{n,\alpha}}(\mathbf{y}^{n,\beta})=\mathbf{y}^{n,\alpha}.$

For M\"obius non-linear activation,
we first map the features $\mathbf{x}^{n,\alpha}\in\mathbb{D}^{n,\alpha}$ into the tangent space $\mathcal{T}_\mathbf{0}\mathbb{D}^{n,\alpha}$ via logarithmic map $\exp_{\mathbf{0}}^\alpha(\cdot)$:
$\log_{\mathbf{0}}^\alpha(\mathbf{x}^{n,\alpha})=\tanh^{-1}(\sqrt{\alpha}\|\mathbf{x}^{n,\alpha}\|)\frac{\mathbf{x}^{n,\alpha}}{\sqrt{\alpha}\|\mathbf{x}^{n,\alpha}\|}$.
Let $l=\tanh^{-1}($ $\sqrt{\alpha}\|\mathbf{x}^{n,\alpha}\|)/({\sqrt{\alpha}\|\mathbf{x}^{n,\alpha}\|})$,
we have $\log_{\mathbf{0}}^\alpha(\mathbf{x}^{n,\alpha})=l\mathbf{x}^{n,\alpha}$.
Also, note that the LeaklyRelu activation function satisfies:
$\sigma(\log_{\mathbf{0}}^\alpha(\mathbf{x}^{n,\alpha}))=l\sigma(\mathbf{x}^{n,\alpha})$.
M\"obius pointwise non-linear activation in Eq. \eqref{eq_pr_m_nonl} is equivalent to:
\begin{equation}
	\begin{split}
		\exp_{\mathbf{0}}^\alpha\!\!\big(\!l\sigma\!(\!\mathbf{x}^{n,\alpha}\!)\!\big)
		\!=\!\frac{1}{\sqrt{\alpha}}\tanh\!\left(\! \frac{\|\!\sigma(\mathbf{x}^{n,\alpha})\!\|}{\|\!\mathbf{x}^{n,\alpha}\!\|}\tanh^{-1}\!\big(\!\sqrt{\alpha}\|\!\mathbf{x}^{n,\alpha}\!\|\!\big) \!\right)\!\cdot\!\frac{\sigma(\mathbf{x}^{n,\alpha})}{\|\sigma(\mathbf{x}^{n,\alpha})\|}.
	\end{split}
\end{equation}
Moreover,
for Lorentzian pointwise non-linear activation,
similar to Eq. \eqref{eq_prove_log_x},
we also map the feature $\mathbf{x}^{n,\beta}=(x_0^{(\beta)}, x_1^{{(\beta)}}, \cdots, x_n^{(\beta)})$ to the tangent space $\mathcal{T}_\mathbf{0}\mathbb{D}^{n,\beta}$ via $\log_{\mathbf{0}}^{\beta}$:
\begin{equation}
	\log_{\mathbf{0}}^\beta(\mathbf{x}^{n,\beta})=\sqrt{\beta}\cosh^{-1}\big(\frac{x_0^{(\beta)}}{\sqrt{\beta}}\big)\frac{(0,\hat{\mathbf{x}}^{(\beta)})}{\|\hat{\mathbf{x}}^{(\beta)}\|}
	=q(0,\hat{\mathbf{x}}^{(\beta)}),
\end{equation}
where 
$\hat{\mathbf{x}}^{(\beta)}=(x_1^{(\beta)},\cdots,x_n^{(\beta)})$, and 
$q=\sqrt{\beta}\cosh^{-1}(x_0^{\beta}/\sqrt{\beta})/\|\hat{\mathbf{x}}^{(\beta)}\!\|$.
Thus, the results of Eq. \eqref{eq_prove_l_non2}
for LeaklyRelu is given as:
$\hat\sigma^{\otimes^\beta}\!\!\big(\!\!\log_{\mathbf{0}}^\beta\!(\mathbf{x}^{n,\beta}\!)\!\big)$
$=q\big( 0,  \sigma(\hat{\mathbf{x}}^{(\beta)})\big)=\mathbf{m}$.
Also the Lorentzian norm of $\mathbf{m}$ also satisfies as:
$\|\mathbf{m}\|_\mathcal{L}=\|q\sigma(\hat{\mathbf{x}})\|=q\|\sigma(\hat{\mathbf{x}})\|$.
The Lorentzian  non-linear activation is given as:
\begin{equation}
	\begin{split}
		\sigma^{\otimes^\beta}\!\!(\mathbf{x}^{n,\beta})
		\!=\!\Big(\! \sqrt{\beta} \cosh\big(\frac{\|\mathbf{m}\|_\mathcal{L}}{\sqrt{\beta}}\big),\frac{\sqrt{\beta}\sinh\big(\frac{\|\mathbf{m}\|_\mathcal{L}}{\sqrt{\beta}}\big)q}{\|\mathbf{m}\|_\mathcal{L}}\cdot \sigma(\mathbf{\hat{x}}^{(\beta)}) \!\Big).
	\end{split}
\end{equation}
Furthermore,
according to Eq. \eqref{eq_poincare2lorentz},
we project the $\sigma^{\otimes^\beta}(\mathbf{x}^{n,\beta})=\mathbf{y}^{n,\beta}$ into the Poincar\'e ball model as following:
$
p_{\mathbb{H}^{n,\beta}\rightarrow \mathbb{D}^{n,\alpha}}(\mathbf{y}^{n,\beta})
=\frac{\sinh\big(\frac{\|\mathbf{m}\|_\mathcal{L}}{\sqrt{\beta}}\big)}{1+\cosh\big( \frac{\|\mathbf{m}\|_\mathcal{L}}{\sqrt{\beta}} \big)}\cdot\!\frac{\sqrt{\beta}}{\|\mathbf{m}\|_\mathcal{L}}\cdot q\sigma\!(\!\mathbf{\hat{x}}^{(\beta)}\!)
\!\!=\!\!\frac{1}{\sqrt{\alpha}}\!\tanh\!\left(\! \frac{\|\sigma(\mathbf{x}^{n,\alpha})\|}{\|\mathbf{x}^{n,\alpha}\|}\!\cdot  \! \tanh^{\!\!-1}\!(\!\sqrt{\alpha}\|\!\mathbf{x}^{n,\alpha}\!\|)  \!\right)\cdot\frac{\sigma(\mathbf{x}^{n,\alpha})}{\|\sigma(\mathbf{x}^{n,\alpha})\|}
=\mathbf{y}^{n,\alpha}.$

Therefore, Lorentzian pointwise non-linear activation is equivalence to M\"obius pointwise non-linear activation.$\hfill\square$ 

\section{Experimental details}

\subsection{$\delta$-hyperbolicity}\label{sec:app_hyperbolicity}
Here we introduce the hyperbolicity measurements originally proposed by Gromov \cite{gromov1987hyperbolic}. 
Considering a quadruple of distinct nodes  $v_1, v_2, v_3, v_4$ in a graph $G$, 
and let $\pi=(\pi_1, \pi_2, \pi_3, \pi_4)$ be a rearrangement of $1, 2, 3, 4$,
which satisfies:
$
S_{\!v_1, v_2, v_3, v_4}\!\!\!=\!\!d^G\!(\!v_{\!\pi_1},\! v_{\!\pi_2}\!) + d^G\!(\!v_{\!\pi_3}, v_{\!\pi_4}\!)
\leq M_{v_1, v_2, v_3, v_4}\!=\!d^G\!(v_{\pi_1}, v_{\pi_3}\!) + d^G(v_{\pi_2}, v_{\pi_4})
\leq L_{v_1, v_2, v_3, v_4}\!=\!d^G\!(v_{\pi_1}, v_{\pi_4}\!) + d^G(v_{\pi_2}, v_{\pi_3}),
$
where $d^G(\cdot,\cdot)$ denotes the graph distance, i.e., the shortest path length,
and let $\delta_{v_1, v_2, v_3, v_4}^+\!\!=\!({L_{v_1, v_2, v_3, v_4}\!-\!M_{v_1, v_2, v_3, v_4}})/{2}$.
The worst case hyperbolicity is to find four nodes in graph $G$ to maximize $\delta_{v_1, v_2, v_3, v_4}^+$, i.e.,
$\delta_{worst}(G)=\max_{v_1, v_2, v_3, v_4}{\delta_{v_1, v_2, v_3, v_4}^+},$
and the average hyperbolicity is to average all combinations of four nodes as:
$\delta_{avg}(G)=\frac{1}{\binom{|V|}{4}}\sum_{v_1, v_2, v_3, v_4}{\delta_{v_1, v_2, v_3, v_4}^+},$
where $|V|$ indicates the number of nodes in the graph.
Note that $\delta_{worst}$,
used in \cite{chami2019hyperbolic},
is a worst case measurement,
which focuses on a local quadruple nodes,
and does not reflect the hyperbolicity of the whole graph \cite{tifrea2018poincar}.
Moreover,
both time complexity of $\delta_{worst}$ and $\delta_{avg}$ are $O(|V|^4)$.
Since $\delta_{avg}$ is robust to adding/removing an edge from the graph,
it can be approximated via sampling,
while $\delta_{worst}$ cannot.
Therefore we leverage $\delta_{avg}$ as the measurement.

\end{document}